\documentclass{article}

\usepackage{microtype}
\usepackage{graphicx}
\usepackage{subfigure}
\usepackage{booktabs} 
\usepackage{hyperref}

\usepackage[accepted]{icml2024}

\usepackage{amsmath}
\usepackage{amssymb}
\usepackage{mathtools}
\usepackage{amsthm}
\usepackage{multirow}
\usepackage{float}
\usepackage[capitalize,noabbrev]{cleveref}

\usepackage[section]{placeins}
\usepackage{multirow}
\usepackage{bm}
\usepackage{colortbl}

\theoremstyle{plain}
\newtheorem{theorem}{Theorem}[section]
\newtheorem{proposition}[theorem]{Proposition}

\newtheorem{corollary}[theorem]{Corollary}
\theoremstyle{definition}

\theoremstyle{remark}

\usepackage[textsize=tiny]{todonotes}

\begin{document}

\twocolumn[
\icmltitle{Predictive Dynamic Fusion}




\begin{icmlauthorlist}
\icmlauthor{Bing Cao}{cic_tju,key}
\icmlauthor{Yinan Xia}{cic_tju}
\icmlauthor{Yi Ding}{cic_tju}
\icmlauthor{Changqing Zhang}{cic_tju,key}
\icmlauthor{Qinghua Hu}{cic_tju,key}
\end{icmlauthorlist}

\icmlaffiliation{cic_tju}{College of Intelligence and Computing, Tianjin University, Tianjin, China}

\icmlaffiliation{key}{Tianjin Key Lab of Machine Learning, Tianjin, China}

\icmlcorrespondingauthor{Changqing Zhang}{zhangchangqing@tju.edu.cn}
\icmlcorrespondingauthor{Qinghua Hu}{huqinghua@tju.edu.cn}

\icmlkeywords{Machine Learning, ICML}

\vskip 0.3in
]



\printAffiliationsAndNotice{}  

\begin{abstract}
Multimodal fusion is crucial in joint decision-making systems for rendering holistic judgments. Since multimodal data changes in open environments, dynamic fusion has emerged and achieved remarkable progress in numerous applications. However, most existing dynamic multimodal fusion methods lack theoretical guarantees and easily fall into suboptimal problems, yielding unreliability and instability. To address this issue, we propose a Predictive Dynamic Fusion (PDF) framework for multimodal learning. We proceed to reveal the multimodal fusion from a generalization perspective and theoretically derive the predictable Collaborative Belief (Co-Belief) with Mono- and Holo-Confidence, which provably reduces the upper bound of generalization error. Accordingly, we further propose a relative calibration strategy to calibrate the predicted Co-Belief for potential uncertainty. Extensive experiments on multiple benchmarks confirm our superiority. Our code is available at \url{https://github.com/Yinan-Xia/PDF}.
\end{abstract}

\section{Introduction}
Many decision-making challenges in real-world applications, such as autonomous driving \cite{cui2019multimodal,feng2020deep}, clinical diagnosis \cite{perrin2009multimodal,tempany2015multimodal}, and sentiment analysis \cite{soleymani2017survey,zadeh2017tensor}, are fundamentally based on multimodal data \cite{kiela2019supervised}. 
To fully capture complementary perceptions, multimodal fusion emerges as a promising learning paradigm that presents an opportunity to integrate all available modalities and achieve enhanced performance. Despite these advances, experiments have shown that traditional fusion techniques have largely overlooked the dynamically changing quality of multimodal data \cite{natarajan2012multimodal,perez2019mfas,yan2004learning}. In reality, the data quality of different modalities and their inherent relationships often vary with the open environment. Numerous studies \cite{xue2023dynamic} empirically recognized that multimodal learning sometimes falls to depending on partial modalities, even a single modality, instead of multimodal data, especially with modality imbalance \cite{wang2020makes,peng2022balanced} or high noise \cite{huang2021learning,scheunders2007wavelet}. Therefore, dynamic multimodal learning becomes a key cue for robust fusion. Some recent works theoretically proved multimodal learning models do not always outperform their unimodal counterparts, encountering limited data volumes \cite{huang2021makes}. This indicates that the dynamic relationship between multimodal data is not a free lunch.

\begin{figure}[t]
\vskip 0.2in
\begin{center}
\begin{subfigure}
  \centering
  \includegraphics[width=0.45\textwidth]{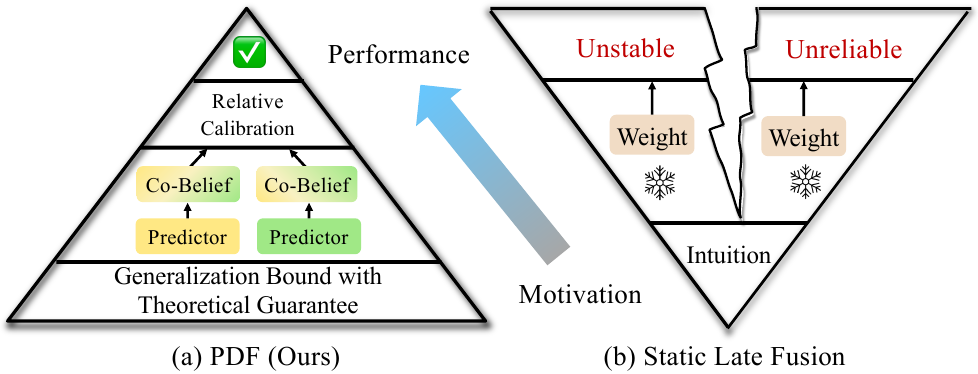}
  \label{fig:sub1}
\end{subfigure}
\begin{subfigure}
  \centering
  \includegraphics[width=0.45\textwidth]{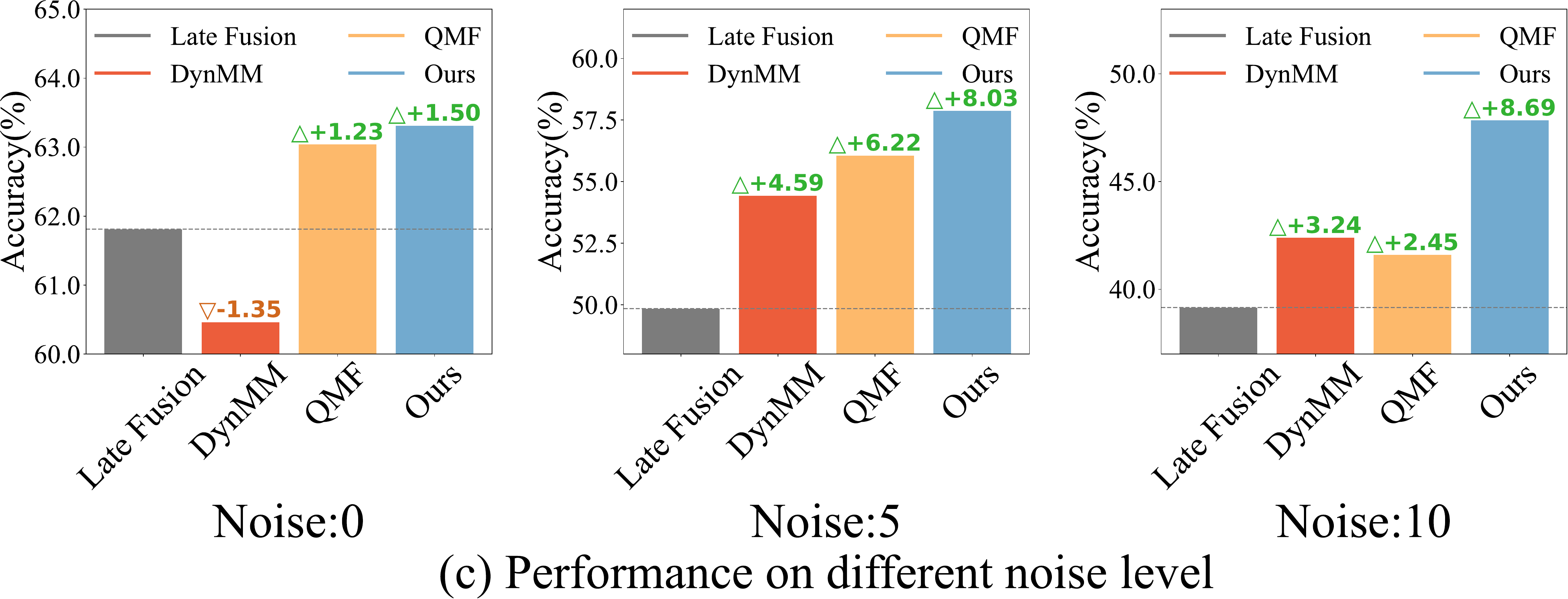}
  \label{fig:sub2}
\end{subfigure}
\vspace{-7pt}
\caption{Our PDF \textit{v.s.} other fusion methods. We derive from the upper bound of generalization error and predict the Co-Belief for each modality with a theoretical guarantee. The relative calibration calibrates potential uncertainty for more reliable learning. Experiments on different noise levels validate our superiority.}
\label{figure: coverfigure_combined}
\vspace{-8pt}
\end{center}
\vskip -0.2in
\end{figure}

Intuitively, fusing information from multimodal data by using the overall quality estimation of each modality is reasonable. However, the data quality estimation is not always reliable due to the unimodal uncertainty and the changing relative reliability of multimodal systems \cite{ma2023calibrating}. We empirically identify that the dominance of each modality is dynamically changing in open environments. On this basis, one fundamental challenge to reliable multimodal learning is how to precisely estimate the contribution of each modality to the multimodal systems \cite{10.5555/3618408.3620161}. However, existing multimodal dynamic fusion techniques mainly focus on addressing this problem by exploring dynamic network architecture \cite{xue2023dynamic} or estimating the modality's quality by uncertainty \cite{han2022trusted}, which commonly lack theoretical guarantees, resulting in unsatisfactory fusion performance.

To solve this problem, we revisited the relationship between modality fusion weights and losses. Deriving from the upper bound of generalization error (Theorem 3.5 in \citealt{mohri2018foundations}), we reveal that the key to reducing generalization error bound lies in the negative covariance between fusion weight and current modality loss as well as the positive covariance between fusion weight and other's modality loss, which implies that fusion weights in the multimodal system should not solely consider the unimodal but is also enforced to integrate other modalities' statuses. With this finding, a natural idea is to employ the loss value of each modality to perform multimodal fusion. However, directly predicting the loss value is unstable as the loss is minimized when converged (see \cref{sec:tcp_loss}). In the setting of multimodal classification with cross-entropy loss, we transform the prediction of loss value into the confidence of true class label, while satisfying the correlation derived from the generalization error. The motivation is based on a natural intuition, i.e., the probability of true class and loss is negative.

To this end, we offer a new perspective on the theoretical foundation of multimodal fusion and propose a Predictive Dynamic Fusion (PDF) framework, which is effective in reducing the upper bound of generalization error and significantly improving multimodal reliability and stability. As shown in \cref{figure: coverfigure_combined} (a), to be specific, PDF predicts the Collaborative Belief (Co-Belief) with Mono-Confidence and Holo-Confidence for each modality. The Mono- and Holo-Confidences derive from the intra-modal negative and inter-modal positive covariance between fusion weight and loss function, respectively. In addition, we empirically identify the changing data quality in open environments, which leads to inevitable prediction uncertainty. To handle this issue, we further propose a relative calibration to calibrate the predicted Co-Belief from the perspective of a multimodal system, which implies that the relative dominance of each modality should change dynamically as the quality of other modalities changes, rather than being static. Experiments demonstrated that our method has strong generalization capabilities, achieving superior results on multiple datasets. Overall, our contributions can be summarized as follows:

\label{submission}
\vspace{-12pt}
\begin{itemize}
    \item This paper provides an intuitive and rigorous multimodal fusion paradigm from the perspective of generalization error. Under theoretical analysis, we derive a new Predictive Dynamic Fusion (PDF) framework based on the covariance of the fusion weight and loss function. This offers theoretical guarantees to reduce the upper bound of generalization error in decision-level multimodal fusion.
    \vspace{-4pt}
    \item We propose to transform the loss prediction to a more robust Collaborative Belief (Co-Belief) prediction, which naturally satisfies the covariance relationship to reduce the upper bound of generalization error without additional computational cost, and significantly enhance the prediction stability.
    \vspace{-4pt}
    \item We develop a relative calibration strategy to calibrate the potential prediction uncertainty and reveal the relative dominance in dynamic multimodal systems. Extensive experiments validate our theoretical analysis and superior performance.
\end{itemize}

\section{Related Works}
\textbf{Multimodal fusion}
is a fundamental problem in multimodal learning \cite{atrey2010multimodal,cao2023multi,zhu2024task}.
Existing methods can be mainly categorized into early fusion \cite{nefian2002dynamic}, middle fusion \cite{natarajan2012multimodal}, and late fusion \cite{snoek2005early,wang2019multi}. Early fusion \cite {ayache2007classifier} directly combines various modalities at the data level, often merging multimodal data through concatenation. Middle fusion \cite{han2022multimodal,wang2019gmc} is widely used in multimodal learning, which mainly fuses multimodal data at the feature level. Late fusion \cite{10.5555/3618408.3620161} usually integrates multimodal data in the semantic space, which can be further grouped into naive fusion \cite{liu2018late}, learnable classifier fusion \cite{xue2023dynamic}, and confidence-based fusion \cite{han2022trusted}. 

\textbf{Uncertainty estimation}
is crucial for improving the model's interpretability, accuracy, and robustness, especially for multimodal systems. Many efforts \cite{neal2012bayesian,gal2016dropout} have been made on this issue. Bayesian Neural Networks (BNNs) \cite{denker1990transforming,mackay1992bayesian} use probability distributions, rather than single values, to represent the weights in neural networks. Deep ensemble methods \cite{lakshminarayanan2017simple,amini2020deep} typically train multiple models and aggregate their predictions, 
then estimate the uncertainty through prediction variances.
Dempster-Shafer's theory extends Bayesian to subjective probabilities, offering a robust model for handling epistemic uncertainty~\cite{dempster1968generalization}. 
Energy Score~\cite{liu2020energy} is promising in estimating uncertainty. 
The pioneering work, QMF~\cite{10.5555/3618408.3620161} explores generalization error and uncertainty-aware weighting to perform robust fusion.
Gradient-based uncertainty~\cite{lee2020gradients} uses backward propagation gradients to quantify uncertainty. Essentially, it evaluates the output uniformity.

\begin{figure*}[t]
\vskip 0.2in
\begin{center}
\centerline{\includegraphics[width=0.8\textwidth]{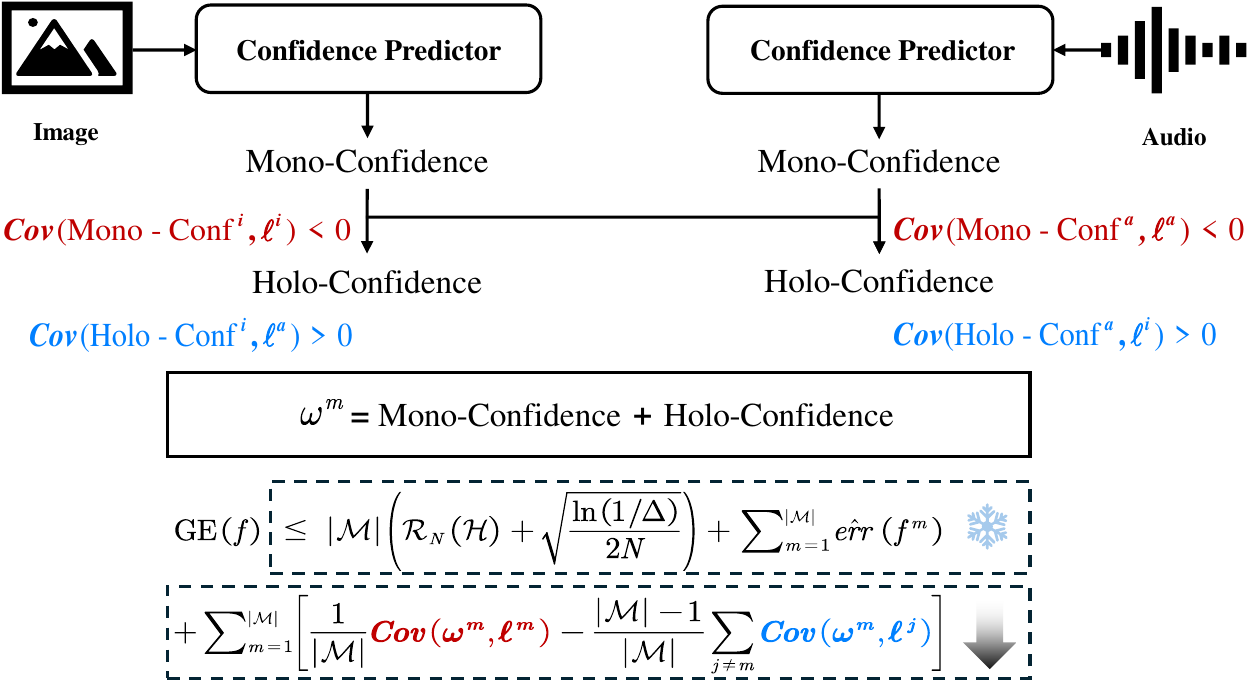}}
\vspace{-3pt}
\caption{
We use confidence predictors to predict the Mono-Confidence of each modality, where the confidence is negatively correlated with the loss of the corresponding modality theoretically. Taking into account the Mono-Confidence of other modalities, we further obtain the Holo-Confidence, where the confidence is positively correlated with the loss of other modalities. By combining Mono-Confidence and Holo-Confidence, we obtain the Co-Belief, which is calibrated as fusion weight to achieve a reduction in the generalization error bounds.} \label{figure: frame}
\vspace{-15pt}
\end{center}
\vskip -0.1in
\end{figure*}

\section{Theory} \label{sec: theory}
In this section, we first clarify the basic settings and formulas in multimodal fusion. Next, we revisit the formula for generalization error bounds and establish its connection with fusion weights, revealing the theoretical guarantee for reducing the upper bound of generalization error. Finally, we propose a predictable dynamic fusion framework that satisfies the above theoretical analysis.

\subsection{Basic Setting}
Given multimodal tasks, we define $\mathcal{M}$ as the set of modalities, thus $\lvert\mathcal{M}\rvert$ is the cardinality of $\mathcal{M}$. We denote our training dataset as $\mathcal{D}_{train}=\{x_i,y_i\}_{i=1}^N\in\mathcal{X}\times\mathcal{Y}$, where $N$ is the sample size of $\mathcal{D}_{train}$, $x_i=\{x_i^{m}\}_{m=1}^{\lvert \mathcal{M}\rvert}$ has $\lvert\mathcal{M}\rvert$ modalities, and $y_i\in\mathcal{Y}$ is the corresponding label. We aim to design a predictable fusion weight $\omega$ for each modality and achieve a robust multimodal fusion. 
The uni-modal projection function $f^m:\mathcal{X}\xrightarrow{}\mathcal{Y}$ is trained as the fusion weight $\omega^m$ dynamically adjusting during training, where $m\in\mathcal{M}$. The decision-level multimodal fusion is as:
\begin{equation}
f(x)=\sum_{m=1}^{\lvert\mathcal{M}\rvert}\omega^m\cdot f^m(x^{m}). \label{equ:fusion}
\end{equation}

\subsection{Generalization Error Upper Bound} \label{sec:3.2}
Generalization Error Upper Bound (GEB) is an important concept in machine learning, referring to an upper bound on the performance of a model on unknown data~\cite{10.5555/3618408.3620161}. Typically, the smaller the upper bound of the generalization error, the better the model's generalization ability, i.e., the better it performs on unknown joint distribution. For binary classification, the Generalization Error (GE) of a model $f$ can be defined as:
\begin{align}
    \text{GE}(f)=\mathbb{E}_{(x,y)\sim\mathcal{D}}[\ell(f(x),y)],
\end{align}
where $\ell$ is a convex logistic loss function, $\mathcal{D}$ is an unknown dataset. By Rademacher complexity theory (Theorem 3.5 in \citealt{mohri2018foundations}), we delve into the essence of GEB in multimodal fusion and obtain the following theorem. The full proof of \cref{thm:GEB} is given in \cref{app:proof_GEB}.

\begin{theorem}\label{thm:GEB} (Generalization Error Upper Bound in Multimodal System).
Let $\hat{err}(f^m)$ denotes the empirical errors of the $m$-th modality on $\mathcal{D}_{train}=\{x_i,y_i\}_{i=1}^N$, and $\mathcal{H}$ is hypothesis set i.e., $\mathcal{H}:\mathcal{X}\rightarrow\{-1,+1\}$, where $f\in \mathcal{H}$. $\mathcal{R}_N(\mathcal{H})$ is the Rademacher complexities (Theorem 3.5 in \citealt{mohri2018foundations}). We holds with a confidence level of $1-\Delta$ ($0<\Delta<1$):
    \begin{align}
    &\mathrm{GE}(f)\leq\lvert\mathcal{M}\rvert\left(\mathcal{R}_N(\mathcal{H})+\sqrt{\frac{\ln(1/\Delta)}{2N}}\right)+\sum\limits_{m=1}\limits^{\lvert\mathcal{M}\rvert}\hat{err}(f^m)\notag\\
&+\sum\limits_{m=1}\limits^{\lvert\mathcal{M}\rvert}\bigg[\frac{1}{\lvert\mathcal{M}\rvert}\underbrace{Cov(\omega^m,\ell^m)}_{\text{\textcolor[RGB]{192,0,0}{Mono-Covariance}}}-\frac{\lvert\mathcal{M}\rvert-1}{\lvert\mathcal{M}\rvert}\sum_{j\neq m}\underbrace{{Cov(\omega^m,\ell^j)}}_{\text{\textcolor[RGB]{0,128,255}{Holo-Covariance}}}\bigg],
     \label{equ:cov}
\end{align}
\end{theorem}
where $Cov(\omega^m, \ell^m)$ is the covariance between fusion weight and loss of $m$-th modality, and $Cov(\omega^m, \ell^j)$ is the cross-modal covariance.
Note that the empirical errors $\hat{err}(f^m)$ and the Rademacher complexities $\mathcal{R}_N(\mathcal{H})$ remain constant when optimizing over the same model. Therefore, the key to achieving a lower GEB lies in ensuring that $Cov(\omega^m, \ell^m)<0$ and $Cov(\omega^m, \ell^j)>0, \forall j\neq m$. Thus, to reduce the multimodal fusion model's GEB, we can draw the following corollaries:
\begin{corollary} \label{coro:1}
A negative correlation should exist between a modality's weight and its loss.
\end{corollary}
\begin{corollary} \label{coro:2}
A positive correlation is desirable between a modality's weight and the losses of the other modalities.
\end{corollary}

\subsection{Collaborative Belief}\label{sec:CB}

\subsubsection{Mono-Confidence}\label{sec:monoconfidence}
To fulfill \cref{coro:1}, an intuitive strategy is to predict the loss of each modality and utilize the predicted loss to formulate the weight, thereby establishing a negative correlation explicitly and directly. Nevertheless, employing loss as a fusion weight for modalities presents significant challenges. Notably, as the loss minimizes during the training process, even marginal biases can induce substantial perturbations. This sensitivity to small errors in loss estimation may compromise the stability and effectiveness of the weight.
Also, the loss value may range from zero to positive infinity, making its precise prediction quite challenging. To mitigate these challenges, we propose substituting loss with the probability ($p_{true}\in[0,1]$) of the true class label, which is inversely related to the loss as denoted by $\ell=-\log p_{true}$ (Full derivation of the relationship between loss and $p_{true}$ is provided in the \cref{app:tcp_loss}), provides a more stable and interpretable basis for weight computation.

Owing to the negative correlation between loss and $p_{true}$, we consider using the $p_{true}$ as the weight for multimodal fusion to fulfill \cref{coro:1}. By analyzing the properties of $p_{true}$, we identify that it reflects the confidence of modality, some works \cite{corbiere2019addressing} have articulated this.
Using $p_{true}$ as a fusion weight not only helps in lowering the upper bound of generalization error but also provides a theoretical guarantee for dynamic multimodal fusion. Since the predictable $p_{true}$ solely considers the current modality's confidence, we define it as Mono-Confidence: 
\begin{align}
    \text{Mono-Conf}^m=\hat{p}_{true}^m, \label{equ:w1}
\end{align}
where $\hat{p}^m_{true}$ is the prediction of $p^m_{true}$ as there is no ground-true label in the test phase. The detailed implementation of the prediction is given in \cref{sec:loss}
\subsubsection{Holo-Confidence} \label{sec:holo}
Recalling \cref{coro:2}, an instinctive approach to naturally achieve $Cov(\omega^m,\ell^j)>0,\forall j\neq m$ is using the losses of other modalities as the weight. Thus, we consider constructing the weight by using the sum of the losses from other modalities as the weight for the given modality. Basing the property of $\ell=-\log p_{true}$, we replaced $\ell$ with $p_{true}$. Thus, we define this term as Holo-Confidence because of the cross-modal interaction of $p_{true}$:
\begin{align}
    \text{Holo-Conf}^m&=\frac{\sum_{j\neq m}^{\lvert\mathcal{M}\rvert}\hat{\ell}^j}{\sum_{i=1}^{\lvert\mathcal{M}\rvert}\hat{\ell}^i}\notag\\
    &=\frac{\log\prod_{j\neq m}\hat{p}_{ture}^j}{\log\prod_{i=1}^{\lvert\mathcal{M}\rvert}\hat{p}_{true}^i} \label{equ:inter},
\end{align}
where $\hat{\ell}^i$ and $\hat{\ell}^j$ is the prediction of $\ell^i$ and $\ell^j$. Our proposed Holo-Confidence also fulfills \cref{coro:2}, the full proof is given in the \cref{app:LC}.

\subsubsection{Co-Belief} \label{sec:co-confidence}
Since Mono-Confidence and Holo-Confidence facilitate the collaborative interaction among modalities, 
to fulfill \cref{coro:1} and \ref{coro:2} simultaneously, we define a collaborative belief (Co-Belief) as a linear combination of the predictable Mono-Confidence and Holo-Confidence, which can be taken as the final fusion weight.
\begin{equation} \label{eq:co-conf}
    \text{Co-Belief}^m=\text{Mono-Conf}^m+\text{Holo-Conf}^m.
\end{equation}
Noting that the Co-Belief meets \cref{coro:1} and \ref{coro:2} simultaneously, and is better than the individual term to represent the weight. The proof is shown in \cref{app:coconf}.

\section{Method}\label{sec:pdf}
To achieve a reliable prediction, we propose a relative calibration strategy to calibrate the predicted Co-Belief to handle inevitable uncertainty. With this reliable prediction, we coined our multimodal fusion framework as \textbf{P}redictive \textbf{D}ynamic \textbf{F}usion (PDF).

Note that the data quality usually dynamically changes in open environments, leading to inevitable uncertainty for the prediction. To decrease the potential uncertainty of Co-Belief in complex scenarios, we further propose a Relative Calibration (RC) to calibrate the predicted Co-Belief from the perspective of a multimodal system. This implies that the relative dominance of each modality should change dynamically as the quality of other modalities changes, rather than being static.

Firstly, we define the \textbf{D}istribution \textbf{U}niformity $\mathrm{DU}^m$ of $m$-th modality in multimodal system as,
\begin{align} \label{equ:DB}
    \mathrm{DU}^m=\frac{1}{C}\sum\limits_{i=1}\limits^{C}\lvert Softmax(f^m(x^{m}))_i-\mu\rvert,
\end{align}
where $C$ is the class number, $\mu$ is the mean of probability, and it holds $\mu = \frac{1}{C}$. 
The distribution of probabilities after softmax offers critical insights into a model's uncertainty: A uniform distribution typically suggests high uncertainty, whereas a peaked distribution implies low uncertainty in predictions \cite{huang2021importance}. We compared it with other uncertainty estimation methods in \cref{app: uncertainty}.

Considering the changing environment, the uncertainty of different modalities in a multimodal system should be relative, i.e., the uncertainty of each modality should change dynamically as the uncertainty of other modalities changes. One modality should dynamically perceive the changes of other modalities and modify its relative contribution to the multimodal system. Thus, we introduce a \textbf{R}elative \textbf{C}alibration (RC) to calibrate the relative uncertainty of each modality.
The relative calibration for $m$-th modality can be formulated as follows (in a scenario with two modalities, denoted as \( m, n \in \mathcal{M} \)): 
\begin{align} 
    \mathrm{RC}^m=\frac{\mathrm{DU}^m}{\mathrm{DU}^n}=\frac{\sum\limits_{i=1}\limits^{C}\lvert Softmax(f^m(x^{m}))_i-\mu\rvert}{\sum\limits_{i=1}\limits^{C}\lvert Softmax(f^n(x^{n}))_i-\mu\rvert}.\label{equ:grb}
\end{align}
The definition of $\mathrm{RC}^m$ when $\lvert\mathcal{M}\rvert>2$ is given in the \cref{app:m>3}.

Considering real-world factors, $\mathrm{RC}^m$ works with an asymmetric form to further calibrate the Co-Belief. Specifically, we postulate that the modality with a $\mathrm{RC}^m<1$ possess greater uncertainty,
and it tends to produce relatively unreliable predictions for $\hat{p}_{true}^m$ \cite{gawlikowski2023survey}, 
thus the corresponding Co-Belief has potential risks in accuracy. Hence, we reduce the contribution of such a modality by multiplying its predicted Co-Belief with $\mathrm{RC}^m$ ($\mathrm{RC}^m<1$). Conversely, modalities with a $\mathrm{RC}^m>1$ are deemed to have less uncertainty and accurate Co-Belief, thus the contribution of these modalities can be maintained to reduce optimization difficulty. Based on this, the asymmetric calibration term is defined as:
\begin{align} \label{eq:k}
    k^m=
    \begin{cases}
        \mathrm{RC}^m
        =\frac{\mathrm{DU}^m}{\mathrm{DU}^n}
         &\text{if } \mathrm{DU}^m<\mathrm{DU}^n,\\
        1 &\text{otherwise.}
    \end{cases}
\end{align}
We calibrate the Co-Belief of the $m$-th modality with our asymmetric calibration strategy and obtain the Calibrated Co-Belief (CCB) as:
\begin{align} \label{eq:cali}
    \text{CCB}^m&=(\text{Co-Belief}^m)\cdot k^m.
\end{align}
We use the each modality's CCB as its fusion weight in a multimodal system,
\begin{align}f(x)&=\sum\limits_{m=1}\limits^{\lvert\mathcal{M}\rvert}\omega^m\cdot f^m(x^{m})\notag\\&=\sum\limits_{m=1}\limits^{\lvert\mathcal{M}\rvert}Softmax(\text{CCB}^m)\cdot f^m(x^{m}).
\end{align}

The loss functions of our PDF framework are given in  \cref{sec:loss}.

\section{Experiments}
\subsection{Setup}
\textbf{Datasets.} We evaluate the proposed method across various multimodal classification tasks, including 
$\diamond$ \textit{Image-text classification}: The UPMC FOOD101 dataset 
\cite{wang2015recipe} contains noisy images and texts obtained in uncontrolled environments containing about 100,000 recipes for a total of 101 food categories. MVSA \cite{niu2016sentiment} is a sentiment analysis dataset that collects sentiment data for matched pairs of users' texts and images; 
$\diamond$ \textit{Scenes recognition}: NYU Depth V2 \cite{silberman2012indoor} is an indoor scenes dataset, both the RGB and Depth Cameras recorded the image-pairs; 
$\diamond$ \textit{Emotion recognition}: CREMA-D \cite{cao2014crema} is an audio-visual dataset designed for recognizing multi-modal emotion, demonstrating various basic emotional states (happy, sad, anger, fear, disgust, and neutral) through spoken sentences. 
$\diamond$ \textit{Face recognition}: PIE \cite{Sim-2003-8822} is a pose, illumination, and expression database of over 40,000 facial images of 68 people.
\begin{table*}[t] 
\renewcommand\arraystretch{1}
\caption{We add Gaussian noise on $50\%$ modalities and $\epsilon$ presents the noise degree. This table reported the average and worst classification accuracies of our method and the competing methods on MVSA, FOOD101, NYU Depth V2, and CREMA-D datasets. The method marked with * was replicated by ourselves, while the rest of the results are sourced from \cite{10.5555/3618408.3620161}. Full results with standard deviation are reported in \cref{tab: Gaussian}.}\label{table:avg_worst}
\vskip 0.2in
\begin{center}
\begin{small}
\begin{sc}
\begin{tabular}{c|ccccccc}
\toprule
& & \multicolumn{2}{c}{$\epsilon$ = $0.0$} & \multicolumn{2}{c}{$\epsilon$ = $5.0$} & \multicolumn{2}{c}{$\epsilon$ = $10.0$} \\
\multirow{-2}{*}{Dataset} & \multirow{-2}{*}{Method} & Avg. & Worst. & Avg. & Worst. & Avg. & Worst. \\ \midrule
& Img & $64.12$ & $62.04$ & $49.36$ & $45.67$ & $45.00$ & $39.31$ \\
& Text  & $75.61$ & $74.76$ & $69.50$ & $65.70$ & $47.41$ & $45.86$ \\
& Concat & $65.59$ & $64.74$ & $50.70$ & $44.70$ & $46.12$ & $41.81$ \\
& Late fusion & $76.88$ & $74.76$ & $63.46$ & $58.57$ & $55.16$ & $47.78$ \\
& TMC & $74.88$ & $71.10$ & $66.72$ & $60.12$ & $60.36$ & $53.37$ \\
& QMF & $78.07$ & $76.30$ & $73.85$ & $71.10$ & $61.28$ & $57.61$ \\
& DynMM* & $79.07$ & $78.23$ & $67.96$ & $65.51$ & $59.21$ & $56.65$ \\ \cline{2-8} 
\multirow{-9}{*}{MVSA}                                                         \rule{0pt}{11pt} & Ours                    
& {$\mathbf{79.94}$} & {$\mathbf{78.42}$} & {$\mathbf{74.40}$} & {$\mathbf{72.64}$} & {$\mathbf{63.09}$} & {$\mathbf{60.31}$} \\ \midrule
& Img & $64.62$ & $64.22$ & $34.72$ & $34.19$ & $33.03$ & $32.67$ \\
& Text & $86.46$ & $86.42$ & $67.38$ & $67.19$ & $43.88$ & $43.56$ \\
& Concat & $88.20$ & $87.81$ & $61.10$ & $59.25$ & $49.86$ & $47.79$ \\
& Late fusion & $90.69$ & $90.58$ & $68.49$ & $65.05$ & $58.00$ & $55.77$ \\
& TMC & $89.86$ & $89.80$ & $73.92$ & $73.64$ & $61.37$ & $61.10$ \\
& QMF & $92.92$ & $92.72$ & $76.03$ & $74.68$ & $62.21$ & $61.76$ \\
& DynMM* & $92.59$ & $92.50$ & $74.74$ & $74.35$ & $59.68$ & $59.22$\\ \cline{2-8} 
\multirow{-9}{*}{\begin{tabular}[c]{@{}c@{}}UMPC \\FOOD 101\end{tabular}}                                                     \rule{0pt}{11pt}& Ours                     & {$\mathbf{93.32}$} & {$\mathbf{92.84}$} & {$\mathbf{76.47}$} & {$\mathbf{76.09}$} & {$\mathbf{62.83}$} & {$\mathbf{62.03}$} \\ \midrule
& RGB & $63.30$ & $62.54$ & $53.12$ & $50.31$ & $45.46$ & $42.20$\\
& Depth & $62.65$ & $61.01$ & $50.95$ & $42.81$ & $44.13$ & $35.93$ \\
& Concat* & $69.88$ & $69.11$ & $63.82$ & $61.47$ & $60.03$ & $55.66$ \\
& Late fusion* & $70.03$ & $68.65$ & $64.37$ & $63.30$ & $60.55$ & $57.95$ \\
& TMC*  & $70.40$ & $70.03$ & $59.33$ & $55.51$ & $50.61$ & $45.41$ \\
& QMF* & $69.54$ & $68.65$ & $64.10$ & $62.54$ & $60.18$ & $58.41$ \\
& DynMM* & $65.50$ & $64.99$ & $54.31$ & $52.14$ & $46.79$ & $45.26$ \\ \cline{2-8} 
\multirow{-9}{*}{\begin{tabular}[c]{@{}c@{}}NYU \\      Depth V2\end{tabular}}\rule{0pt}{11pt} & Ours & {$\mathbf{71.37}$} & {$\mathbf{70.18}$} & {$\mathbf{65.72}$} & {$\mathbf{63.91}$} & {$\mathbf{62.56}$} & {$\mathbf{60.25}$} \\ \midrule
& Visual* & $43.60$ & $40.05$ & $32.52$ & $28.49$ & $30.17$ & $28.09$ \\
& Audio* & $58.67$ & $57.39$ & $54.66$ & $50.67$ & $43.01$ & $35.35$ \\
& Concat* & $61.56$ & $59.95$ & $52.33$ & $45.16$ & $41.01$ & $31.59$ \\
& Late fusion* & $61.81$ & $57.39$ & $49.84$ & $39.92$ & $39.15$ & $29.90$ \\
& TMC* & $59.15$ & $56.18$ & $54.42$ & $45.16$ & $46.79$ & $37.63$ \\
& QMF* & $63.04$ & $60.75$ & $56.06$ & $51.75$ & $41.60$ & $35.89$\\
& DynMM* & $60.46$ & $59.81$ & $54.43$ & $52.82$ & $42.39$ & $41.26$ \\ \cline{2-8} 
\multirow{-8}{*}{CREMA-D}                                                     \rule{0pt}{11pt} & Ours & {$\mathbf{63.31}$} & {$\mathbf{61.69}$} & {$\mathbf{57.85}$} & {$\mathbf{54.17}$} & {$\mathbf{47.84}$} & {$\mathbf{44.62}$} \\ \bottomrule
\end{tabular}
\end{sc}
\end{small}
\end{center}
\vskip -0.1in
\end{table*}

\textbf{Evaluation metrics.} We report the average and worst accuracies in the presence of Gaussian noise (for image and audio modalities) and blank noise (for text modality), in accordance with prior studies \cite{10.5555/3618408.3620161,han2022trusted,xie2017data,ma2021trustworthy}. To mitigate the impact of randomness, we replicate to evaluate our model using five distinct seeds. We also defined two new metrics, \textbf{A}ggregate \textbf{C}ovariance (AC) and \textbf{G}EB \textbf{D}ecreasing \textbf{P}roportion (GDP), to quantify the capability of fusion strategy to reduce generalization error upper bound, i.e. the 
generalization ability of the model with certain fusion strategy. The specific definitions are given in \cref{app:metric}.

\textbf{Competing methods.} 
In our experiments, we compare our method with established fusion techniques, including late fusion and concatenation-based fusion, which are static, as well as with dynamic fusion approaches, including TMC \cite{han2022trusted}, QMF \cite{10.5555/3618408.3620161} and DynMM \cite{xue2023dynamic}. 
We also establish unimodal baselines for comparison: RGB and depth for scene recognition; text and image for image-text classification; visual and audio for emotion recognition.

\textbf{Implementation details.}
The network was trained for 100 epochs utilizing the Adam optimizer with $\beta_1=0.9$, $\beta_2=0.999$, weight decay of 0.01, dropout rate of 0.1, and a batch size of 16. 
All the experiments were conducted on an NVIDIA A6000 GPU, using PyTorch with default parameters for all methods. More details are provided in \cref{app:implement}.

\begin{figure*}[th]
\vskip 0.2in
\begin{center}
\centerline{\includegraphics[width=1\textwidth]{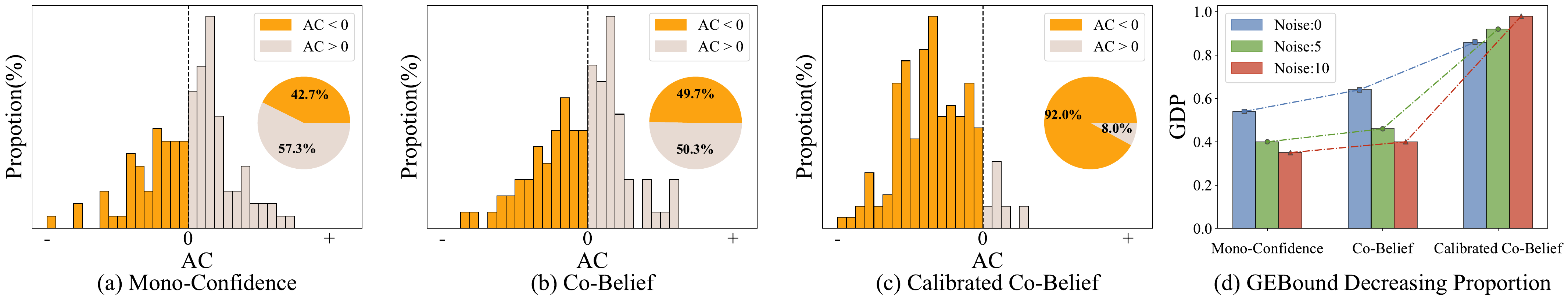}}
\vspace{-3pt}
\caption{
We evaluated the effectiveness of Mono-Confidence, Co-Belief, and Calibrated Co-Belief as fusion weights on the NYU Depth V2 dataset to determine their effectiveness in minimizing the generalization error upper bound.
The yellow part of the pie chart in \cref{figure: GDP} (a), (b), or (c) illustrates the Generalization Error Bound Decreasing Proportion (GDP) for each weight form under varying noises (0, 5, and 10).
As depicted in \cref{figure: GDP} (c), the Calibrated Co-Belief attains the highest GDP, leading to the best generalization. \cref{figure: GDP} (d) presents the GDP across diverse fusion strategies and noise intensities. More details are given in \cref{app:lower GEB}.
} \label{figure: GDP}
\vspace{-7pt}
\end{center}
\vskip -0.2in
\end{figure*}

\vspace{-2pt}
\subsection{Questions to be Verified}
We conducted a series of experiments to investigate our effectiveness and rationale thoroughly. The experiments were designed to address four primary questions:

\begin{itemize}
    \vspace{-3pt}
    \item \textit{Does our proposed method have better generalization ability than its counterparts?}
    
    In \cref{sec:3.2}, we conducted a theoretical analysis to demonstrate that our method can effectively lower the upper bound of generalization error, which is evidenced in its performance and robustness to noisy data.
    To empirically substantiate that our method reduces the generalization error upper bound, we carried out comparative experiments across five datasets under diverse noise conditions.
    \vspace{-3pt}
    \item \textit{Does our PDF framework really work?}

    We performed an ablation study to verify the effectiveness of each component of our framework. Additionally, we visualized the capability of Mono-Confidence, Co-Belief, and Calibrated Co-Belief in reducing the generalization error upper bound.
    \item  \textit{Why do we predict $p_{true}$ instead of loss?}
    
    In \cref{sec:CB}, we choose to establish the correlation between weights and loss by predicting $p_{true}$. We analyzed the distribution of $p_{true}$, and by examining the relationship between $p_{true}$ and loss, we identified the challenge in predicting loss. Eventually, we compared the performance of the two prediction methods, validating the advantages of $p_{true}$.
    \vspace{-3pt}
    \item \textit{Why relative calibration is effective and reliable?}
    
    To further investigate the effect of relative calibration, we conducted experiments to explore the relative uncertainty when data quality changes by adding noise. We also compared the efficacy of DU with that of other uncertainty estimation methods.
\end{itemize}

\subsection{Results}
\subsubsection{Generalization Ability} 
Our method improves the model's generalization compared to the competing approaches, as shown in  \cref{table:avg_worst}. Notably, as the noise intensity increases, the advantages of our method become increasingly highlighted, reinforcing its superior generalization potential. It is especially commendable that our approach consistently realizes state-of-the-art performance on all the datasets against the competing methods. Full results under different noises (Gaussian noise and Salt-Pepper noise) with standard deviation are shown in \cref{app:full_noise}. Additionally, we also conducted experiments on PIE dataset with 3 modalities in \cref{app: multimodal}.

To further validate the efficacy of our approach, we incorporated time-varying noise into the CREMA-D dataset to emulate real-world scenarios. Specifically, we introduced noise with varying frequencies to synthesize noisy speech data, and the intensity of noise added to image frames also varied over time. As depicted in \cref{tab: timenoise}, our PDF demonstrated exceptional generalization prowess, even amidst the influence of time-varying noise interference.

\begin{table}[t]
\vskip -0.1in
\caption{Ablation study on MVSA to verify the effectiveness of Mono-Confidence (MC), Holo-Confidence (HC), and relative calibration (RC) as well as the complete model.} \label{table:ablation}
\vskip -0.2in
\renewcommand\arraystretch{1.1}
\tabcolsep=0.05cm
\begin{center}
\begin{small}
\begin{sc}
\begin{tabular}{ccccccccc}
\toprule
                        &                         &                       & \multicolumn{2}{c}{$\epsilon$ = 0}                                            & \multicolumn{2}{c}{$\epsilon$ = 5}                                            & \multicolumn{2}{c}{$\epsilon$ = 10}                                           \\
\multirow{-2}{*}{MC} & \multirow{-2}{*}{HC} & \multirow{-2}{*}{RC} & Avg.                                  & Worst.                                & Avg.                                  & Worst.                                & Avg.                                  & Worst.                                \\ \midrule
$\surd$                       &                         &                       & $79.43$                                 & $78.23$                                 &$ 72.57$                                 & $69.56$                                 &$60.84$                                 &$ 55.11$                                 \\
                        & $\surd$                       &                       & $79.28$                                 & $78.23$                                 & $72.57$                                 & $69.94$                                 &$ 61.22$                                 & $55.68$                                 \\
                        &                         & $\surd$                     & $79.11$                                 & $77.84 $                                & $71.30$                                 & $63.97$                                 & $60.11$                                 & $50.29$                                 \\
$\surd$                       & $\surd$                       &                       & $79.92 $                                & $\mathbf{79.00}$ & $72.83$                                 & $70.13 $                                & $60.97$                                 &$ 55.49$                                 \\
                        & $\surd$                       & $\surd$                     &$ 79.06$                                 & $78.23$                                 & $73.09 $                                & $71.10 $                                &$ 62.11$                                 &$ 58.57 $                                \\
$\surd$                       &                         & $\surd$                     &$ 79.62$                                 & $78.23 $                                & $73.12 $                                & $70.91$                                 & $62.13   $                              & $57.61$                                 \\ \midrule
$\surd$                       & $\surd$                       & $\surd$                     & {  $\mathbf{79.94}$} & $78.42 $                                & {  $\mathbf{74.40}$} & {  $\mathbf{72.64}$} & {  $\mathbf{63.09}$} & {  \textbf{60.31}} \\ \bottomrule
\end{tabular}
\end{sc}
\end{small}
\end{center}
\vskip -0.2in
\end{table}
\vskip 0.13in
\subsubsection{Ablation Study}\label{sec:5.2.1}
We conducted ablated experiments on Mono-Confidence, Holo-Confidence, and Relative Calibration. The mean and worst accuracy on the MVSA dataset across different noise intensities are reported. The results, presented in \cref{table:ablation}, reveal that the model with Calibrated Co-Belief attains the best robustness and generalization.

\begin{figure}[t]
\vskip 0.1in
\begin{center}
\centerline{\includegraphics[width=0.5\textwidth]{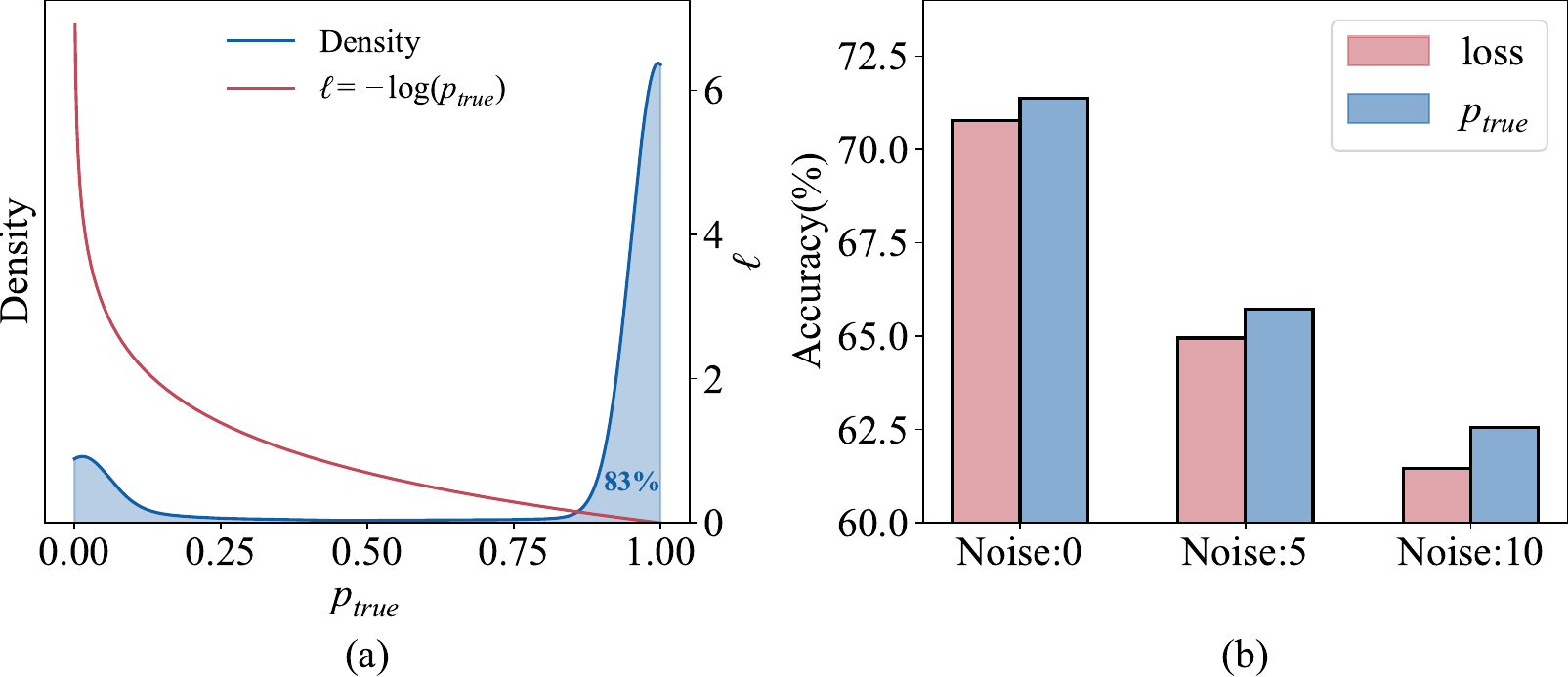}}
\caption{
We present the true distribution of $p_{true}$ for the samples in UPMC Food 101, according to the blue area in (a), while the red line in (a) is the function curve of loss corresponding to $p_{true}$. In (b), we reported the performance of two prediction methods in various noise conditions. It's obvious that predicting $p_{true}$ yields better performance.
} \label{fig:tcp_loss}
\vspace{-12pt}
\end{center}
\vskip -0.1in
\end{figure}
\vspace{-4pt}
\begin{figure}[t]
\vskip 0.1in
\begin{center}
\centerline{\includegraphics[width=0.5\textwidth]{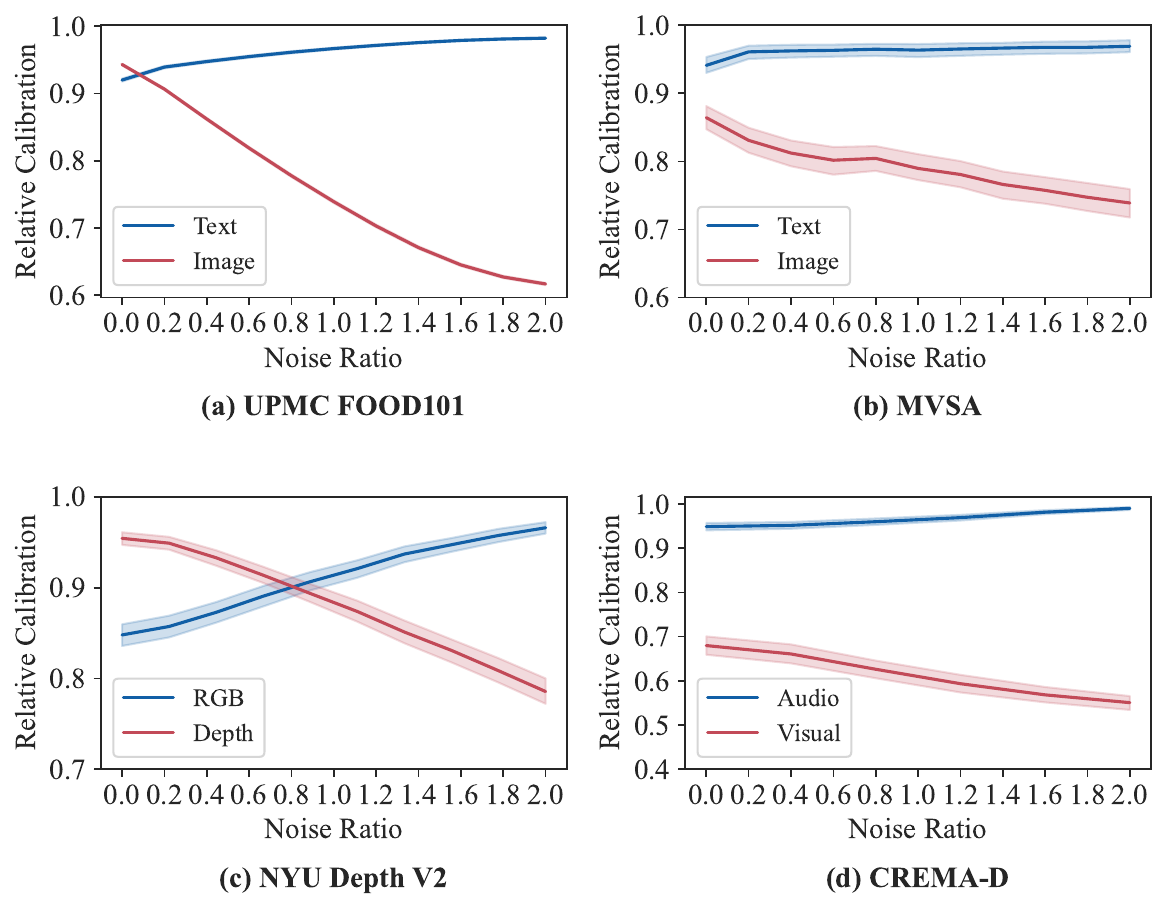}}
\vspace{-9pt}
\caption{
Relative Calibration (RC) can detect noise variations within the current modality as well as in other modalities. 
The noise ratio denotes the ratio of the noises added to the two modalities, representing the relative exposure of the two modalities to noise. We maintained a fixed noise level for the modality denoted by the blue line in the figure.
} \label{figure:dist}
\vspace{-15pt}
\end{center}
\vskip -0.1in
\end{figure}
\vspace{10pt}
\begin{table}[t] 
\vskip 0.1in
\caption{Comparison between our PDF and other competing methods on CREMA-D dataset with added time-varying noise.}
\vskip -0.1in
\begin{center}
\begin{small}
\begin{sc}
\centering
\label{tab: timenoise}
\vskip 0.1in
\renewcommand\arraystretch{1.1}
\tabcolsep=0.5cm
\begin{tabular}{lcc}
\toprule
Method    & Avg. & Worst. \\ 
\midrule 
Concat & $60.20$ & $58.60$ \\
Late Fusion & $60.80$ & $57.39$ \\
TMC & $57.17$  & $54.03$  \\
QMF & $61.20$ & $59.40$\\
DynMM & $57.90$ & $57.53$\\
\midrule
Ours & $\mathbf{61.40}$ & $\mathbf{60.34}$\\
\bottomrule
\end{tabular}
\vspace*{-10pt}
\end{sc}
\end{small}
\end{center}
\end{table}

\textbf{Lower generalization error upper bound.}
We applied the proposed Generalization Error Bound Decreasing Proportion (GDP) to measure the capacity of a fusion strategy for reducing the Generalization Error Upper Bound (GEB). Separately, we compare the GDP metrics of models that utilize ${\text{Mono-Conf}}$ (\cref{equ:w1}), ${\text{Co-Belief}}$ (\cref{equ:inter}), and $\text{CCB}$ (\cref{eq:cali}) as fusion weights. Specifically, for each fusion strategy, we trained 50 models with distinct random seeds and reported the GDP of these models under varying noises (0, 5, and 10) during testing. We took Mono-Confidence, Co-Belief, and CCB as distinct fusion weights for the model and depicted the AC (defined in \cref{eq:goal}) distribution in \cref{figure: GDP} (a), (b), and (c). The proportion of $\text{AC} < 0$, i.e. GDP, signifies the fusion strategy's potential to reduce the model's GEB. \cref{figure: GDP} (d) displays the GDP of different fusion strategies under diverse noisy conditions. The proportion of Mono-Confidence that reduces GEB without noise is greater than 50\%, and Holo-Confidence further increases the model's GDP, CCB endows the model with the best generalization capabilities in dynamic environments. These experiments validate our effectiveness in lowering the generalization error upper bound.

\begin{table}[t] 
\caption{Comparison with traditional uncertainty on MVSA.}\label{tab:uncertainty}
\vskip 0.1in
\renewcommand\arraystretch{1.1}
\tabcolsep=0.35cm
\begin{center}
\begin{small}
\begin{sc}
\begin{tabular}{lccc}
\toprule
\multicolumn{1}{c}{Uncertainty} & $\epsilon$ = 0 & $\epsilon$ = 5 & $\epsilon$ = 10 \\ \midrule
Energy                         & $79.14$          & $73.22$          & $61.72$           \\
Entropy                         & $78.88$          &$ 72.07$          & $60.94$           \\
Evidence                        & $79.17$          &$73.28 $          &$61.87 $ \\
MCP                             & $79.48$          &$ 73.22$          & $61.93$           \\ \midrule
Ours (DU)                    & $\mathbf{79.94}$ & $\mathbf{74.40}$ & $\mathbf{63.09}$  \\ 
\bottomrule
\end{tabular}
\end{sc}
\end{small}
\end{center}
\vskip -0.1in
\end{table}

\subsubsection{\texorpdfstring{Predicting \(p_{true}\) is More Feasible}{Predicting p\_true is More Feasible}}\label{sec:tcp_loss}

Although the most direct method to establish a relationship between weight and loss is to use the predicted loss as weight, it encountered difficulties in practice. As shown in \cref{fig:tcp_loss} (a), we display the sample distribution of $p_{true}$ and the corresponding loss function curve. We observed that approximately 83\% of $p_{true}$ values fall into the range of 0.8 to 1, while the corresponding loss value ranges between 0 and 0.097, making it challenging to predict loss accurately. Employing the same methodology, we conducted comparative experiments on predicting the loss and the $p_{true}$, \cref{fig:tcp_loss} (b) shows that predicting $p_{true}$ consistently outperforms predicting loss under various noise conditions. This experiment validates the effectiveness of our prediction strategy.

\subsubsection{Effectiveness of Relative Calibration}
\textbf{Relative calibration reflects the quality of modality.}
We conducted experiments across four datasets with two modalities to explore the responsiveness of Relative Calibration (RC) to modality quality. We changed the data quality by varying noise levels. Specifically, we alter the degree of noise added to it and fix another modality's noise level. More details of adding noise are given in \cref{sec:noise details}.

The noise ratio denotes the ratio of the noises added to the two modalities, representing the relative exposure of the two modalities to noise. As illustrated in \cref{figure:dist}, the RC value of the noisy modality declines with increased noise, while the RC of the modality with fixed noise enhances or maintains as the noise level of the other modality escalates. This indicates that RC is adept at discerning the quality of both its modality and the others. These findings corroborate the conceptual expectation of multimodal data quality relativity and emphasize the dynamism and interpretability of RC.

\textbf{Comparison with traditional uncertainty.}
We compared the proposed $\mathrm{DU}$ with conventional uncertainty estimation methods. For a fair comparison, we substituted our proposed DU in RC with alternative uncertainties, namely energy-based uncertainty \cite{liu2020energy}, entropy \cite{shannon1948mathematical}, uncertainty in Dempster–Shafer Theory (DST) \cite{sensoy2018evidential}, and Maximum Class Probability (MCP) \cite{hendrycks2016baseline}. As shown in \cref{tab:uncertainty}, DU demonstrates the best performance compared to other uncertainties. Among these uncertainties, entropy has a similar form to DU, however, it is unsuitable for our method, more details are given in \cref{app: uncertainty}.
Moreover, to validate the advantages of RC's relativity and asymmetric form, we conducted analysis and experiments in \cref{app:RR}.

\section{Conclusion}
Through extensive empirical studies, we observe that the fusion paradigms of existing methods are typically unreliable, and lack theoretical guarantees. Starting from the generalization error upper bound (GEB), we find the positive and negative correlations between fusion weight and loss, which inspired us to predict the Mono- and Holo-Confidence directly. Thus, we obtain predictable Co-Belief with theoretical guarantees to reduce GEB. Due to the potential prediction uncertainty, it is further calibrated in multimodal systems by relative calibration and used as the fusion weight. Comprehensive experiments with in-depth analysis validate our superiority in accuracy and stability against other approaches. Moreover, our PDF's extensions to other tasks are worth exploring. We believe this method is inspirational research that will benefit the community.

\section*{Impact Statement}
This paper presents work to advance the field of multimodal fusion in machine learning. Our goal is to construct a predictive multimodal fusion method to boost the safety and accuracy of joint decisions in multimodal systems, lowering the potential modality bias and instability of prediction. However, due to the modality imbalance and data bias in open environments, there is a possibility of inevitable uncertainty when applying our method in real-world applications.

\section*{Acknowledgements}
This work was supported in part by the National Natural Science Foundation of China under Grants 62106171, 61925602, and 62376193, in part by the Tianjin Natural Science Foundation under Grant 21JCYBJC00580, and in part by the Key Laboratory of Big Data Intelligent Computing, Chongqing University of Posts and Telecommunications under Grant BDIC-2023-A-008. This work was also sponsored by CAAI-CANN Open Fund, developed on OpenI Community. 
\textit{Yinan Xia} and \textit{Yi Ding} contributed equally to this work.


\begin{thebibliography}{57}
\providecommand{\natexlab}[1]{#1}
\providecommand{\url}[1]{\texttt{#1}}
\expandafter\ifx\csname urlstyle\endcsname\relax
  \providecommand{\doi}[1]{doi: #1}\else
  \providecommand{\doi}{doi: \begingroup \urlstyle{rm}\Url}\fi

\bibitem[Amini et~al.(2020)Amini, Schwarting, Soleimany, and Rus]{amini2020deep}
Amini, A., Schwarting, W., Soleimany, A., and Rus, D.
\newblock Deep evidential regression.
\newblock In \emph{Advances in Neural Information Processing Systems}, volume~33, pp.\  14927--14937, 2020.

\bibitem[Atrey et~al.(2010)Atrey, Hossain, El~Saddik, and Kankanhalli]{atrey2010multimodal}
Atrey, P.~K., Hossain, M.~A., El~Saddik, A., and Kankanhalli, M.~S.
\newblock Multimodal fusion for multimedia analysis: a survey.
\newblock \emph{Multimedia Systems}, 16:\penalty0 345--379, 2010.

\bibitem[Ayache et~al.(2007)Ayache, Qu{\'e}not, and Gensel]{ayache2007classifier}
Ayache, S., Qu{\'e}not, G., and Gensel, J.
\newblock Classifier fusion for svm-based multimedia semantic indexing.
\newblock In \emph{European Conference on Information Retrieval}, pp.\  494--504. Springer, 2007.

\bibitem[Cao et~al.(2023)Cao, Sun, Zhu, and Hu]{cao2023multi}
Cao, B., Sun, Y., Zhu, P., and Hu, Q.
\newblock Multi-modal gated mixture of local-to-global experts for dynamic image fusion.
\newblock In \emph{Proceedings of the IEEE/CVF International Conference on Computer Vision}, pp.\  23555--23564, 2023.

\bibitem[Cao et~al.(2014)Cao, Cooper, Keutmann, Gur, Nenkova, and Verma]{cao2014crema}
Cao, H., Cooper, D.~G., Keutmann, M.~K., Gur, R.~C., Nenkova, A., and Verma, R.
\newblock Crema-d: Crowd-sourced emotional multimodal actors dataset.
\newblock \emph{IEEE Transactions on Affective Computing}, 5\penalty0 (4):\penalty0 377--390, 2014.

\bibitem[Corbi{\`e}re et~al.(2019)Corbi{\`e}re, Thome, Bar-Hen, Cord, and P{\'e}rez]{corbiere2019addressing}
Corbi{\`e}re, C., Thome, N., Bar-Hen, A., Cord, M., and P{\'e}rez, P.
\newblock Addressing failure prediction by learning model confidence.
\newblock In \emph{Advances in Neural Information Processing Systems}, volume~32, 2019.

\bibitem[Cui et~al.(2019)Cui, Radosavljevic, Chou, Lin, Nguyen, Huang, Schneider, and Djuric]{cui2019multimodal}
Cui, H., Radosavljevic, V., Chou, F.-C., Lin, T.-H., Nguyen, T., Huang, T.-K., Schneider, J., and Djuric, N.
\newblock Multimodal trajectory predictions for autonomous driving using deep convolutional networks.
\newblock In \emph{2019 International Conference on Robotics and Automation (ICRA)}, pp.\  2090--2096. IEEE, 2019.

\bibitem[Dempster(1968)]{dempster1968generalization}
Dempster, A.~P.
\newblock A generalization of bayesian inference.
\newblock \emph{Journal of the Royal Statistical Society: Series B (Methodological)}, 30\penalty0 (2):\penalty0 205--232, 1968.

\bibitem[Denker \& LeCun(1990)Denker and LeCun]{denker1990transforming}
Denker, J. and LeCun, Y.
\newblock Transforming neural-net output levels to probability distributions.
\newblock In \emph{Advances in Neural Information Processing Systems}, volume~3, 1990.

\bibitem[DeVries \& Taylor(2018)DeVries and Taylor]{devries2018learning}
DeVries, T. and Taylor, G.~W.
\newblock Learning confidence for out-of-distribution detection in neural networks.
\newblock \emph{arXiv preprint arXiv:1802.04865}, 2018.

\bibitem[Feng et~al.(2020)Feng, Haase-Sch{\"u}tz, Rosenbaum, Hertlein, Glaeser, Timm, Wiesbeck, and Dietmayer]{feng2020deep}
Feng, D., Haase-Sch{\"u}tz, C., Rosenbaum, L., Hertlein, H., Glaeser, C., Timm, F., Wiesbeck, W., and Dietmayer, K.
\newblock Deep multi-modal object detection and semantic segmentation for autonomous driving: Datasets, methods, and challenges.
\newblock \emph{IEEE Transactions on Intelligent Transportation Systems}, 22\penalty0 (3):\penalty0 1341--1360, 2020.

\bibitem[Gal \& Ghahramani(2016)Gal and Ghahramani]{gal2016dropout}
Gal, Y. and Ghahramani, Z.
\newblock Dropout as a bayesian approximation: Representing model uncertainty in deep learning.
\newblock In \emph{International Conference on Machine Learning}, pp.\  1050--1059. PMLR, 2016.

\bibitem[Gawlikowski et~al.(2023)Gawlikowski, Tassi, Ali, Lee, Humt, Feng, Kruspe, Triebel, Jung, Roscher, et~al.]{gawlikowski2023survey}
Gawlikowski, J., Tassi, C. R.~N., Ali, M., Lee, J., Humt, M., Feng, J., Kruspe, A., Triebel, R., Jung, P., Roscher, R., et~al.
\newblock A survey of uncertainty in deep neural networks.
\newblock \emph{Artificial Intelligence Review}, 56\penalty0 (Suppl 1):\penalty0 1513--1589, 2023.

\bibitem[Han et~al.(2022{\natexlab{a}})Han, Yang, Huang, Zhang, and Yao]{han2022multimodal}
Han, Z., Yang, F., Huang, J., Zhang, C., and Yao, J.
\newblock Multimodal dynamics: Dynamical fusion for trustworthy multimodal classification.
\newblock In \emph{Proceedings of the IEEE/CVF Conference on Computer Vision and Pattern Recognition}, pp.\  20707--20717, 2022{\natexlab{a}}.

\bibitem[Han et~al.(2022{\natexlab{b}})Han, Zhang, Fu, and Zhou]{han2022trusted}
Han, Z., Zhang, C., Fu, H., and Zhou, J.~T.
\newblock Trusted multi-view classification with dynamic evidential fusion.
\newblock \emph{IEEE Transactions on Pattern Analysis and Machine Intelligence}, 45\penalty0 (2):\penalty0 2551--2566, 2022{\natexlab{b}}.

\bibitem[Hendrycks \& Gimpel(2016)Hendrycks and Gimpel]{hendrycks2016baseline}
Hendrycks, D. and Gimpel, K.
\newblock A baseline for detecting misclassified and out-of-distribution examples in neural networks.
\newblock \emph{arXiv preprint arXiv:1610.02136}, 2016.

\bibitem[Huang et~al.(2021{\natexlab{a}})Huang, Geng, and Li]{huang2021importance}
Huang, R., Geng, A., and Li, Y.
\newblock On the importance of gradients for detecting distributional shifts in the wild.
\newblock In \emph{Advances in Neural Information Processing Systems}, volume~34, pp.\  677--689, 2021{\natexlab{a}}.

\bibitem[Huang et~al.(2021{\natexlab{b}})Huang, Du, Xue, Chen, Zhao, and Huang]{huang2021makes}
Huang, Y., Du, C., Xue, Z., Chen, X., Zhao, H., and Huang, L.
\newblock What makes multi-modal learning better than single (provably).
\newblock In \emph{Advances in Neural Information Processing Systems}, volume~34, pp.\  10944--10956, 2021{\natexlab{b}}.

\bibitem[Huang et~al.(2021{\natexlab{c}})Huang, Niu, Liu, Ding, Xiao, Wu, and Peng]{huang2021learning}
Huang, Z., Niu, G., Liu, X., Ding, W., Xiao, X., Wu, H., and Peng, X.
\newblock Learning with noisy correspondence for cross-modal matching.
\newblock In \emph{Advances in Neural Information Processing Systems}, volume~34, pp.\  29406--29419, 2021{\natexlab{c}}.

\bibitem[Kiela et~al.(2019)Kiela, Bhooshan, Firooz, Perez, and Testuggine]{kiela2019supervised}
Kiela, D., Bhooshan, S., Firooz, H., Perez, E., and Testuggine, D.
\newblock Supervised multimodal bitransformers for classifying images and text.
\newblock \emph{arXiv preprint arXiv:1909.02950}, 2019.

\bibitem[Lakshminarayanan et~al.(2017)Lakshminarayanan, Pritzel, and Blundell]{lakshminarayanan2017simple}
Lakshminarayanan, B., Pritzel, A., and Blundell, C.
\newblock Simple and scalable predictive uncertainty estimation using deep ensembles.
\newblock In \emph{Advances in Neural Information Processing Systems}, volume~30, 2017.

\bibitem[Lee \& AlRegib(2020)Lee and AlRegib]{lee2020gradients}
Lee, J. and AlRegib, G.
\newblock Gradients as a measure of uncertainty in neural networks.
\newblock In \emph{2020 IEEE International Conference on Image Processing (ICIP)}, pp.\  2416--2420. IEEE, 2020.

\bibitem[Liang et~al.(2017)Liang, Li, and Srikant]{liang2017enhancing}
Liang, S., Li, Y., and Srikant, R.
\newblock Enhancing the reliability of out-of-distribution image detection in neural networks.
\newblock \emph{arXiv preprint arXiv:1706.02690}, 2017.

\bibitem[Liu et~al.(2020)Liu, Wang, Owens, and Li]{liu2020energy}
Liu, W., Wang, X., Owens, J.~D., and Li, Y.
\newblock Energy-based out-of-distribution detection.
\newblock In \emph{Proceedings of the 34th International Conference on Neural Information Processing Systems}, pp.\  21464--21475, 2020.

\bibitem[Liu et~al.(2018)Liu, Zhu, Li, Wang, Tang, Yin, Shen, Wang, and Gao]{liu2018late}
Liu, X., Zhu, X., Li, M., Wang, L., Tang, C., Yin, J., Shen, D., Wang, H., and Gao, W.
\newblock Late fusion incomplete multi-view clustering.
\newblock \emph{IEEE Transactions on Pattern Analysis and Machine Intelligence}, 41\penalty0 (10):\penalty0 2410--2423, 2018.

\bibitem[Ma et~al.(2021)Ma, Han, Zhang, Fu, Zhou, and Hu]{ma2021trustworthy}
Ma, H., Han, Z., Zhang, C., Fu, H., Zhou, J.~T., and Hu, Q.
\newblock Trustworthy multimodal regression with mixture of normal-inverse gamma distributions.
\newblock In \emph{Advances in Neural Information Processing Systems}, volume~34, pp.\  6881--6893, 2021.

\bibitem[Ma et~al.(2023)Ma, Zhang, Zhang, Wu, Fu, Zhou, and Hu]{ma2023calibrating}
Ma, H., Zhang, Q., Zhang, C., Wu, B., Fu, H., Zhou, J.~T., and Hu, Q.
\newblock Calibrating multimodal learning.
\newblock In \emph{International Conference on Machine Learning}, pp.\  23429--23450. PMLR, 2023.

\bibitem[Mackay(1992)]{mackay1992bayesian}
Mackay, D. J.~C.
\newblock \emph{Bayesian methods for adaptive models}.
\newblock California Institute of Technology, 1992.

\bibitem[Mohri et~al.(2018)Mohri, Rostamizadeh, and Talwalkar]{mohri2018foundations}
Mohri, M., Rostamizadeh, A., and Talwalkar, A.
\newblock \emph{Foundations of machine learning}.
\newblock MIT press, 2018.

\bibitem[M{\"u}ller et~al.(2019)M{\"u}ller, Kornblith, and Hinton]{muller2019does}
M{\"u}ller, R., Kornblith, S., and Hinton, G.~E.
\newblock When does label smoothing help?
\newblock In \emph{Advances in Neural Information Processing Systems}, volume~32, 2019.

\bibitem[Natarajan et~al.(2012)Natarajan, Wu, Vitaladevuni, Zhuang, Tsakalidis, Park, Prasad, and Natarajan]{natarajan2012multimodal}
Natarajan, P., Wu, S., Vitaladevuni, S., Zhuang, X., Tsakalidis, S., Park, U., Prasad, R., and Natarajan, P.
\newblock Multimodal feature fusion for robust event detection in web videos.
\newblock In \emph{2012 IEEE Conference on Computer Vision and Pattern Recognition}, pp.\  1298--1305. IEEE, 2012.

\bibitem[Neal(2012)]{neal2012bayesian}
Neal, R.~M.
\newblock \emph{Bayesian learning for neural networks}, volume 118.
\newblock Springer Science \& Business Media, 2012.

\bibitem[Nefian et~al.(2002)Nefian, Liang, Pi, Liu, and Murphy]{nefian2002dynamic}
Nefian, A.~V., Liang, L., Pi, X., Liu, X., and Murphy, K.
\newblock Dynamic bayesian networks for audio-visual speech recognition.
\newblock \emph{EURASIP Journal on Advances in Signal Processing}, 2002:\penalty0 1--15, 2002.

\bibitem[Niu et~al.(2016)Niu, Zhu, Pang, and El~Saddik]{niu2016sentiment}
Niu, T., Zhu, S., Pang, L., and El~Saddik, A.
\newblock Sentiment analysis on multi-view social data.
\newblock In \emph{MultiMedia Modeling: 22nd International Conference, MMM 2016, Miami, FL, USA, January 4-6, 2016, Proceedings, Part II 22}, pp.\  15--27. Springer, 2016.

\bibitem[Papadopoulos et~al.(2001)Papadopoulos, Edwards, and Murray]{papadopoulos2001confidence}
Papadopoulos, G., Edwards, P.~J., and Murray, A.~F.
\newblock Confidence estimation methods for neural networks: A practical comparison.
\newblock \emph{IEEE Transactions on Neural Networks}, 12\penalty0 (6):\penalty0 1278--1287, 2001.

\bibitem[Peng et~al.(2022)Peng, Wei, Deng, Wang, and Hu]{peng2022balanced}
Peng, X., Wei, Y., Deng, A., Wang, D., and Hu, D.
\newblock Balanced multimodal learning via on-the-fly gradient modulation.
\newblock In \emph{Proceedings of the IEEE/CVF Conference on Computer Vision and Pattern Recognition}, pp.\  8238--8247, 2022.

\bibitem[P{\'e}rez-R{\'u}a et~al.(2019)P{\'e}rez-R{\'u}a, Vielzeuf, Pateux, Baccouche, and Jurie]{perez2019mfas}
P{\'e}rez-R{\'u}a, J.-M., Vielzeuf, V., Pateux, S., Baccouche, M., and Jurie, F.
\newblock Mfas: Multimodal fusion architecture search.
\newblock In \emph{Proceedings of the IEEE/CVF Conference on Computer Vision and Pattern Recognition}, pp.\  6966--6975, 2019.

\bibitem[Perrin et~al.(2009)Perrin, Fagan, and Holtzman]{perrin2009multimodal}
Perrin, R.~J., Fagan, A.~M., and Holtzman, D.~M.
\newblock Multimodal techniques for diagnosis and prognosis of alzheimer's disease.
\newblock \emph{Nature}, 461\penalty0 (7266):\penalty0 916--922, 2009.

\bibitem[Scheunders \& De~Backer(2007)Scheunders and De~Backer]{scheunders2007wavelet}
Scheunders, P. and De~Backer, S.
\newblock Wavelet denoising of multicomponent images using gaussian scale mixture models and a noise-free image as priors.
\newblock \emph{IEEE Transactions on Image Processing}, 16\penalty0 (7):\penalty0 1865--1872, 2007.

\bibitem[Sensoy et~al.(2018)Sensoy, Kaplan, and Kandemir]{sensoy2018evidential}
Sensoy, M., Kaplan, L., and Kandemir, M.
\newblock Evidential deep learning to quantify classification uncertainty.
\newblock In \emph{Advances in Neural Information Processing Systems}, volume~31, 2018.

\bibitem[Shannon(1948)]{shannon1948mathematical}
Shannon, C.~E.
\newblock A mathematical theory of communication.
\newblock \emph{The Bell System Technical Journal}, 27\penalty0 (3):\penalty0 379--423, 1948.

\bibitem[Silberman et~al.(2012)Silberman, Hoiem, Kohli, and Fergus]{silberman2012indoor}
Silberman, N., Hoiem, D., Kohli, P., and Fergus, R.
\newblock Indoor segmentation and support inference from rgbd images.
\newblock In \emph{Computer Vision--ECCV 2012: 12th European Conference on Computer Vision, Florence, Italy, October 7-13, 2012, Proceedings, Part V 12}, pp.\  746--760. Springer, 2012.

\bibitem[Sim et~al.(2003)Sim, Baker, and Bsat]{Sim-2003-8822}
Sim, T., Baker, S., and Bsat, M.
\newblock The cmu pose, illumination, and expression database.
\newblock \emph{IEEE Transactions on Pattern Analysis and Machine Intelligence}, 25\penalty0 (12):\penalty0 1615 -- 1618, December 2003.

\bibitem[Snoek et~al.(2005)Snoek, Worring, and Smeulders]{snoek2005early}
Snoek, C.~G., Worring, M., and Smeulders, A.~W.
\newblock Early versus late fusion in semantic video analysis.
\newblock In \emph{Proceedings of the 13th annual ACM international conference on Multimedia}, pp.\  399--402, 2005.

\bibitem[Soleymani et~al.(2017)Soleymani, Garcia, Jou, Schuller, Chang, and Pantic]{soleymani2017survey}
Soleymani, M., Garcia, D., Jou, B., Schuller, B., Chang, S.-F., and Pantic, M.
\newblock A survey of multimodal sentiment analysis.
\newblock \emph{Image and Vision Computing}, 65:\penalty0 3--14, 2017.

\bibitem[Tempany et~al.(2015)Tempany, Jayender, Kapur, Bueno, Golby, Agar, and Jolesz]{tempany2015multimodal}
Tempany, C.~M., Jayender, J., Kapur, T., Bueno, R., Golby, A., Agar, N., and Jolesz, F.~A.
\newblock Multimodal imaging for improved diagnosis and treatment of cancers.
\newblock \emph{Cancer}, 121\penalty0 (6):\penalty0 817--827, 2015.

\bibitem[Wang et~al.(2019{\natexlab{a}})Wang, Yang, and Liu]{wang2019gmc}
Wang, H., Yang, Y., and Liu, B.
\newblock Gmc: Graph-based multi-view clustering.
\newblock \emph{IEEE Transactions on Knowledge and Data Engineering}, 32\penalty0 (6):\penalty0 1116--1129, 2019{\natexlab{a}}.

\bibitem[Wang et~al.(2019{\natexlab{b}})Wang, Liu, Zhu, Tang, Liu, Hu, Xia, and Yin]{wang2019multi}
Wang, S., Liu, X., Zhu, E., Tang, C., Liu, J., Hu, J., Xia, J., and Yin, J.
\newblock Multi-view clustering via late fusion alignment maximization.
\newblock In \emph{IJCAI}, pp.\  3778--3784, 2019{\natexlab{b}}.

\bibitem[Wang et~al.(2020)Wang, Tran, and Feiszli]{wang2020makes}
Wang, W., Tran, D., and Feiszli, M.
\newblock What makes training multi-modal classification networks hard?
\newblock In \emph{Proceedings of the IEEE/CVF conference on Computer Vision and Pattern Recognition}, pp.\  12695--12705, 2020.

\bibitem[Wang et~al.(2015)Wang, Kumar, Thome, Cord, and Precioso]{wang2015recipe}
Wang, X., Kumar, D., Thome, N., Cord, M., and Precioso, F.
\newblock Recipe recognition with large multimodal food dataset.
\newblock In \emph{2015 IEEE International Conference on Multimedia \& Expo Workshops (ICMEW)}, pp.\  1--6. IEEE, 2015.

\bibitem[Wei et~al.(2022)Wei, Xie, Cheng, Feng, An, and Li]{wei2022mitigating}
Wei, H., Xie, R., Cheng, H., Feng, L., An, B., and Li, Y.
\newblock Mitigating neural network overconfidence with logit normalization.
\newblock In \emph{International Conference on Machine Learning}, pp.\  23631--23644. PMLR, 2022.

\bibitem[Xie et~al.(2017)Xie, Wang, Li, L{\'e}vy, Nie, Jurafsky, and Ng]{xie2017data}
Xie, Z., Wang, S.~I., Li, J., L{\'e}vy, D., Nie, A., Jurafsky, D., and Ng, A.~Y.
\newblock Data noising as smoothing in neural network language models.
\newblock \emph{arXiv preprint arXiv:1703.02573}, 2017.

\bibitem[Xue \& Marculescu(2023)Xue and Marculescu]{xue2023dynamic}
Xue, Z. and Marculescu, R.
\newblock Dynamic multimodal fusion.
\newblock In \emph{Proceedings of the IEEE/CVF Conference on Computer Vision and Pattern Recognition}, pp.\  2574--2583, 2023.

\bibitem[Yan et~al.(2004)Yan, Yang, and Hauptmann]{yan2004learning}
Yan, R., Yang, J., and Hauptmann, A.~G.
\newblock Learning query-class dependent weights in automatic video retrieval.
\newblock In \emph{Proceedings of the 12th annual ACM international conference on Multimedia}, pp.\  548--555, 2004.

\bibitem[Zadeh et~al.(2017)Zadeh, Chen, Poria, Cambria, and Morency]{zadeh2017tensor}
Zadeh, A., Chen, M., Poria, S., Cambria, E., and Morency, L.-P.
\newblock Tensor fusion network for multimodal sentiment analysis.
\newblock \emph{arXiv preprint arXiv:1707.07250}, 2017.

\bibitem[Zhang et~al.(2023)Zhang, Wu, Zhang, Hu, Fu, Zhou, and Peng]{10.5555/3618408.3620161}
Zhang, Q., Wu, H., Zhang, C., Hu, Q., Fu, H., Zhou, J.~T., and Peng, X.
\newblock Provable dynamic fusion for low-quality multimodal data.
\newblock In \emph{International conference on machine learning}, pp.\  41753--41769. PMLR, 2023.

\bibitem[Zhu et~al.(2024)Zhu, Sun, Cao, and Hu]{zhu2024task}
Zhu, P., Sun, Y., Cao, B., and Hu, Q.
\newblock Task-customized mixture of adapters for general image fusion.
\newblock In \emph{Proceedings of the IEEE/CVF Conference on Computer Vision and Pattern Recognition}, 2024.

\end{thebibliography}
\bibliographystyle{icml2024}

\newpage
\appendix
\onecolumn
\section*{Appendix}
\section{Proofs}
\subsection{Proof of \cref{thm:GEB}}\label{app:proof_GEB}
\textit{Proof}. Given the decision-level multimodal fusion formula delineated in \cref{equ:fusion}, consider $\ell$ to be the convex logistic loss function applied to binary classification tasks. The softmax function is utilized to normalize $\omega^m$: $\omega^m=\frac{e^{\omega^m}}{\sum_{j=1}^{\lvert\mathcal{M}\rvert}e^{\omega^j}}$. Considering the property of convex function, we have:
\begin{align}
\ell(f(x),y)&=\ell(\sum\limits_{m=1}\limits^{\lvert\mathcal{M}\rvert}\omega^m f^m(x^{(m)}),y)\leq  \sum\limits_{m=1}\limits^{\lvert\mathcal{M}\rvert}\omega^m\ell(f^m(x^{(m)}),y) \label{equ:w}.
\end{align}
When computing the expectation of \cref{equ:w} and leveraging the properties of expectation, the subsequent equation is satisfied. To simplify notation, $\ell(f^m(x), y)$ can be denoted as $\ell^m$ and $\mathcal{D}$ is an unknown dataset:
\begin{align}
    \mathrm{GE}(f)&=\mathbb E_{(x,y)\sim\mathcal{D}}\ell(f(x),y)\notag\\
    &\leq\sum\limits_{m=1}\limits^{\lvert\mathcal{M}\rvert}\mathbb{E}_{(x,y)\sim\mathcal{D}}[\omega^m\ell^m])\notag\\
    &=\frac{1}{\lvert\mathcal{M}\rvert}\left(\lvert\mathcal{M}\rvert\cdot\sum\limits_{m=1}\limits^{\lvert\mathcal{M}\rvert}\mathbb{E}_{(x,y)\sim\mathcal{D}}[\omega^m\ell^m]\right)\notag\\
    &=\frac{1}{\lvert\mathcal{M}\rvert}\cdot\sum\limits_{m=1}\limits^{\lvert\mathcal{M}\rvert}\left(\mathbb E_{(x,y)\sim\mathcal{D}}[\omega^m\ell^m]+(\lvert\mathcal{M}\rvert-1)\mathbb E_{(x,y)\sim\mathcal{D}}[(1-\sum_{j\neq m}\omega^j)\ell^m]\right)\notag\\
    &=\frac{1}{\lvert\mathcal{M}\rvert}\cdot\sum\limits_{m=1}\limits^{\lvert\mathcal{M}\rvert}\bigg(\mathbb E_{(x,y)\sim\mathcal{D}}[\omega^m]E_{(x,y)\sim\mathcal{D}}[\ell^m]+Cov(\omega^m,\ell^m)\notag\\
    &-(\lvert\mathcal{M}\rvert-1)\cdot\sum_{j\neq m}\left(\mathbb E_{(x,y)\sim\mathcal{D}}[\omega^j]\mathbb E_{(x,y)\sim\mathcal{D}}[\ell^m]+Cov(\omega^j,\ell^m)\right)-(\lvert\mathcal{M}\rvert-1)\cdot\mathbb E_{(x,y)\sim\mathcal{D}}[\ell^m]\bigg)\notag\\
    &=\frac{1}{\lvert\mathcal{M}\rvert}\cdot\sum\limits_{m=1}\limits^{\lvert\mathcal{M}\rvert}\bigg[\mathbb E_{(x,y)\sim\mathcal{D}}[\ell^m]\left(\mathbb E_{(x,y)\sim\mathcal{D}}[\omega^m]+(\lvert\mathcal{M}\rvert-1)\cdot\left[1-\sum_{j\neq m}\mathbb E_{(x,y)\sim\mathcal{D}}[\omega^j]\right]\right)\notag\\
    &+Cov(\omega^m,\ell^m)-(\lvert\mathcal{M}\rvert-1)\sum_{j\neq m}Cov(\omega^j,\ell^m)\bigg]\notag\\
    &=\frac{1}{\lvert\mathcal{M}\rvert}\cdot\sum\limits_{m=1}\limits^{\lvert\mathcal{M}\rvert}\bigg[\mathbb E_{(x,y)\sim\mathcal{D}}[\ell^m]\left(\mathbb E_{(x,y)\sim\mathcal{D}}[\omega^m]+(\lvert\mathcal{M}\rvert-1)\cdot\mathbb E_{(x,y)\sim\mathcal{D}}[\omega^m]\right)\notag\\
    &+Cov(\omega^m,\ell^m)-(\lvert\mathcal{M}\rvert-1)\sum_{j\neq m}Cov(\omega^j,\ell^m)\bigg]\notag\\
    &=\frac{1}{\lvert\mathcal{M}\rvert}\cdot\sum\limits_{m=1}\limits^{\lvert\mathcal{M}\rvert}\bigg[\lvert\mathcal{M}\rvert\cdot\mathbb E_{(x,y)\sim\mathcal{D}}[\ell^m]\mathbb E_{(x,y)\sim\mathcal{D}}[\omega^m]+Cov(\omega^m,\ell^m)-(\lvert\mathcal{M}\rvert-1)\sum_{j\neq m}Cov(\omega^j,\ell^m)\bigg]\notag\\
    &=\sum\limits_{m=1}\limits^{\lvert\mathcal{M}\rvert}\left(\mathbb E_{(x,y)\sim\mathcal{D}}[\ell^m]\mathbb E_{(x,y)\sim\mathcal{D}}[\omega^m]+\frac{1}{\lvert\mathcal{M}\rvert}\cdot\bigg[Cov(\omega^m,\ell^m)-(\lvert\mathcal{M}\rvert-1)\sum_{j\neq m}Cov(\omega^m,\ell^j)\bigg]\right)\notag\\
    &\leq\sum\limits_{m=1}\limits^{\lvert\mathcal{M}\rvert}\left(\mathbb{E}_{(x,y)\sim\mathcal{D}}[\ell^m]\notag+\frac{1}{\lvert\mathcal{M}\rvert}Cov(\omega^m,\ell^m)-\frac{\lvert\mathcal{M}\rvert-1}{\lvert\mathcal{M}\rvert}\sum_{j\neq m}{Cov(\omega^m,\ell^j)}\right)\label{eq:longest}
\end{align}
To simplify \cref{eq:longest}, we invoke Rademacher complexity theory \cite{mohri2018foundations} (Theorem 3.5), which establishes that with a confidence level of $1-\Delta$ where $0<\Delta<1$, the following holds:
\begin{align}
    \mathbb{E}_{(x,y)\sim\mathcal{D}}[\ell^m]\leq\hat{err}[f^m]+\mathcal{R}_N(\mathcal{H})+\sqrt{\frac{\ln(1/\Delta)}{2N}}.
\end{align}
In this context, $\hat{err}(f^m)$ represents the empirical error of the unimodal function $f^m$, and $\mathcal{H}$ denotes the hypothesis set, defined as $\mathcal{H}:\mathcal{X}\rightarrow\{-1,+1\}$, which includes $f$ as a member. The Rademacher complexity is denoted by $\mathcal{R}_N(\mathcal{H})$. Consequently, we assert that with a confidence level of $1-\Delta$, where $0<\Delta<1$, the following relationship is upheld:
    \begin{align}
    \mathrm{GE}(f)\leq\lvert\mathcal{M}\rvert\left(\mathcal{R}_N(\mathcal{H})+\sqrt{\frac{\ln(1/\Delta)}{2N}}\right)+\sum\limits_{m=1}\limits^{\lvert\mathcal{M}\rvert}\hat{err}(f^m)+\sum\limits_{m=1}\limits^{\lvert\mathcal{M}\rvert}\bigg[\frac{1}{\lvert\mathcal{M}\rvert}\underbrace{Cov(\omega^m,\ell^m)}_{\text{\textcolor[RGB]{192,0,0}{Mono-Covariance}}}-\frac{\lvert\mathcal{M}\rvert-1}{\lvert\mathcal{M}\rvert}\sum_{j\neq m}\underbrace{{Cov(\omega^m,\ell^j)}}_{\text{\textcolor[RGB]{0,128,255}{Holo-Covariance}}}\bigg].
    \end{align}
    
\subsection{Proof of Mono-Confidence Satisfies \cref{coro:1}} \label{app:tcp_loss}
\textit{Proof}. In classification tasks with a cross-entropy function, the unimodal loss is defined as: 
\begin{align} \label{equ:celoss} \ell=-\sum_{i=1}^N y_i\log p_i. \end{align}

The one-hot label of the $i$-th class is denoted by ${y}_i$, and $p_i$ represents the predicted probability for the $i$-th class. Under the assumption that the $t$-th class is the correct classification, we have ${y}_t=1$ and ${y}_i=0$ for all $i \neq t$. The probability of true class $p_{true}$ is $p_t$ under the above assumption. Consequently, \cref{equ:celoss} can be simplified to: \begin{align}
\ell&=-{y}_t\log p_t-\sum_{i\neq t}{y}_i\log p_i \notag\\
&=-\log p_t \\
&=-\log p_{true} \label{equ:pr1}
\end{align}
Noting that $p_{true}$ naturally correlates with cross-entropy loss.
To substantiate that $Cov(p_{true}^m,\ell^m)<0$, we put forth the following proposition: \begin{proposition} 
\label{pro:cov}
For any two random variables X and Y, the condition $Cov(X, Y)<0$ is equivalent to that X and Y are inversely correlated, and conversely, $Cov(X, Y)>0$ is equivalent to a positive correlation between them.
\end{proposition}
Recalling \cref{equ:pr1}, we have
\begin{align}
    \frac{\mathrm{d}\ell}{\mathrm{d}p_{true}}=-\frac{1}{p_{true}},
\end{align}
where $p_{true}\in[0,1]$. Hence, we have $\frac{\mathrm{d}\ell}{\mathrm{d}p_{true}}\in(-\infty,-1]$, which means $\ell$ is negatively correlated with $p_{true}$ in the domain of $p_{true}$. With the \cref{pro:cov}, the fact that $Cov(p_{true}^m,\ell^m)<0$ holds. Recalling the  \cref{coro:1}, it's proved that using $p_{true}$ as the Mono-Confidence conforms to reducing generalization error upper bound.

\subsection{Proof of Holo-Confidence Satisfies \cref{coro:2}} \label{app:LC}

\textit{Proof}. Recalling the definition of the $m$-th modality's Holo-Confidence:
\begin{align}
    \text{Holo-Conf}^m=\frac{\sum_{j\neq m}\ell^j}{\sum_{i=1}^{\lvert\mathcal{M}\rvert}\ell^i}.
\end{align}

As for the positive correlation between $\text{Holo-Conf}^m$ and $\ell^j$, the derivative of $\text{Holo-Conf}^m$ with respect to $\ell^j$ for all $j\neq m$ is computed as follows: 
\begin{align}
\frac{\partial\text{Holo-Conf}^m}{\partial\ell^j}=\frac{\sum_{i=1}^{\lvert\mathcal{M}\rvert}\ell^i-\sum_{j\neq m}\ell^j}{{(\sum_{i=1}^{\lvert\mathcal{M}\rvert}\ell^i)}^2}=\frac{\ell^j}{{(\sum_{i=1}^{\lvert\mathcal{M}\rvert}\ell^i)}^2},
\end{align}
where $\ell\in[0,+\infty)$. Consequently, since $\frac{\mathrm{d}\text{Holo-Conf}^m}{\mathrm{d}\ell^j}\in[0,+\infty)$, it is established that $\text{Holo-Conf}^m$ is positively correlated with $\ell^j$ within the domain of $\ell^j$, for all $j\neq m$. Given \cref{pro:cov}, it is validated that $Cov(\text{Holo-Conf}^m,\ell^j)>0$ for all $j\neq m$. Referencing \cref{coro:2}, this demonstrates that our proposed Holo-Confidence metric aligns with the reduction of the generalization error upper bound.
\subsection{Proof of Co-Belief Lowers GEB}\label{app:coconf}

To verify whether a fusion strategy can reduce the model's GEB, we proposed the \textbf{A}ggregate \textbf{C}ovariance (AC), where
\begin{align}
    \text{AC} (f)=\sum\limits_{m=1}\limits^{\lvert\mathcal{M}\rvert}\big(Cov(\omega^m,\ell^m)
    -{(\lvert\mathcal{M}\rvert-1)}\sum_{j\neq m}Cov(\omega^m,\ell^j)\big),
\end{align}
where $f\in\mathcal{H}$ is the multimodal function and $f^m$ is the function of $m$-th modality. When $\text{AC}(f)<0$, it is deemed that the $\text{GEB}$ of $f$ is decreased.
We identify that reducing the generalization upper bound requires ensuring our proposed metric, $\text{AC}(f)<0$. The formulation of the proposed Co-Belief is as follows:
\begin{align}
    \text{Co-Belief}^m=p_{true}^m+\frac{\log\prod_{j\neq m}p_{true}^j}{\log\prod_{i=1}^{\lvert\mathcal{M}\rvert}p_{true}^i}=e^{-\ell^m}+\frac{\sum_{j\neq m}\ell^j}{\sum_{i=1}^{\lvert\mathcal{M}\rvert}\ell^i}.
\end{align}

The desirable result is:
\begin{align}\label{eq:goal}
    \text{AC}(f)=\sum\limits_{m=1}\limits^{\lvert\mathcal{M}\rvert}\left(\underbrace{Cov(\text{Co-Belief}^m,\ell^m)}_{\text{Mono-Covariance}}-(\lvert\mathcal{M}\rvert-1)\sum\limits_{j\neq m}\underbrace{Cov(\text{Co-Belief}^m,\ell^j)}_{\text{Holo-Covariance}}\right)<0.
\end{align}
Now, we consider the Mono-Covariance in \cref{eq:goal}.
\begin{align}
    \frac{\mathrm{d}\text{Co-Belief}^m}{\mathrm{d}\ell^m}=-e^{-\ell^m}-\frac{\sum_{j\neq m}\ell^j}{{(\sum_{i=1}^{\lvert\mathcal{M}\rvert}\ell^i)}^2},
\end{align}
where $\ell\in[0,+\infty)$. Hence, we have $\frac{\mathrm{d}\text{Co-Belief}^m}{\mathrm{d}\ell^m}\in(-\infty,-1], $ which means $\text{Co-Belief}^m$ is negatively correlated with $\ell^m$ in the domain of
$\ell^m$. Recalling \cref{pro:cov}, we have $\text{Mono-Covariance}<0$.

Then, consider the Holo-Covariance in \cref{eq:goal}.
\begin{align}
    \frac{\mathrm{d}\text{Co-Belief}^m}{\mathrm{d}\ell^j}=\frac{\sum_{i=1}^{\lvert\mathcal{M}\rvert\ell^i-\sum_{j\neq m}\ell^j}}{{(\sum_{i=1}^{\lvert\mathcal{M}\rvert}\ell^i)}^2}=\frac{\ell^j}{{(\sum_{i=1}^{\lvert\mathcal{M}\rvert}\ell^i)}^2},
\end{align}
where $\ell\in[0,\infty)$. Therefore, $\frac{\mathrm{d}\text{Co-Belief}^m}{\mathrm{d}\ell^j}\in[0,+\infty)$ indicates that $\text{Co-Belief}^m$ is positively correlated with $\ell^j$ in the domain of $\ell^j,\forall j\neq m$. With the \cref{pro:cov}, we have $\text{Holo-Covariance}>0$. Thus, we achieve our goal in \cref{eq:goal}, reducing the generalization error upper bound.

Furthermore, our proposed Co-Belief surpasses both Mono-Confidence and Holo-Confidence in reducing the generalization error upper bound. Unlike $\text{Mono-Conf}^m$, which exhibits no correlation with $\ell^j$, the Mono-Covariance of $\text{Co-Belief}^m$ is less than that of $\text{Mono-Conf}^m$. Similarly, $\text{Holo-Conf}^m$ has a higher Mono-Covariance than $\text{Co-Belief}^m$. These findings underscore our proposed Co-Belief's marked advantage in diminishing the generalization error upper bound, as corroborated by our ablation studies.
\subsection{The Complete Form of the RC Formula} \label{app:m>3}
Revisiting our proposed relative calibration term as defined in \cref{equ:grb}, it is important to note that this term is initially conceptualized under a two-modality setting. For cases where $\lvert\mathcal{M}\rvert>2$, the relative calibration term is redefined as follows:
\begin{align}
    \mathrm{RC}^m=
       \frac{\mathrm{DU}^m\cdot(\lvert\mathcal{M}\rvert-1)}{\sum_{i\neq m}\mathrm{DU}^i}.
\end{align}
Moreover, when formulating the final calibrated Co-Belief, we perform truncation on the RC value for asymmetric calibration, which is defined by the following formula:
\begin{align}\label{eq:multimodal}
    \mathrm{k}^m=
       \begin{cases}\mathrm{RC}^m=\frac{\mathrm{DU}^m\cdot(\lvert\mathcal{M}\rvert-1)}{\sum_{i\neq m}\mathrm{DU}^i} &\text{if }\mathrm{RC}^m<1,\\
           1 &\text{otherwise}.
       \end{cases}
\end{align}
It is observed that when $\lvert\mathcal{M}\rvert=2$, the definition is congruent with the formulation of RC and $k$ as delineated in \cref{equ:grb} and \cref{eq:k}. Furthermore, this definition is consistent with the theoretical analysis presented in the main body of our paper.

\section{More Analysis}
\subsection{\texorpdfstring{$\hat{p}_{true}$ is Reliable}{p\_true-hat is Reliable}}\label{sec:p_true}
Existing confidence estimation methods \cite{corbiere2019addressing, papadopoulos2001confidence} usually depend on estimating confidence precisely for certain tasks in unimodal scenarios. By contrast, in our proposed method, $p_{true}$ is used to construct dynamic fusion weight to satisfy the theoretical guarantee and reflect the modality dominance in multimodal fusion. Therefore, the reliability of $\hat{p}_{true}$ in our method is reflected in the ability to provide reasonable fusion weights (Mono-Confidence) for each modality, which implies that when different modalities make inconsistent decisions, the fusion weight of the dominant modality should be higher and vice versa, to make a correct decision jointly.

\begin{table}[t] 
\caption{We report the proportion (\%) of fusion results that are correct by different fusion weight when the two modalities made conflicting decision.}
\vskip 0.1in
\label{tab: ptrue}
\begin{center}

\begin{small}
\begin{sc}
\centering

\begin{tabular}{lcc}

\toprule
Method    & MVSA & UMPC FOOD 101 \\ 
\midrule
Late Fusion & $60.58$ & $89.31$ \\
TMC & $60.09$  & $90.84$  \\
QMF & $73.45$ & $93.77$\\
DynMM & $76.77$ & $88.94$\\
\midrule
Ours & $\mathbf{77.49}$ & $\mathbf{95.53}$\\
\bottomrule
\end{tabular}
\vspace{-18pt}
\vskip -0.1in
\end{sc}
\end{small}
\end{center}
\end{table}

To validate the reliability of $\hat{p}_{true}$, We calculated the probability that $\hat{p}_{true}$ is able to help make the right decision in case unimodal models make false classifications. 
For the two-modalities dataset, we first counted the number of samples in which the two modalities made conflicting decisions. Among these conflicting samples, we then calculated the proportion of cases where the $\hat{p}_{true}$-weighted (Mono-Confidence weighted) fusion result was correct. We compare $\hat{p}_{true}$-weighted fusion with other methods that have uni-modal outputs. As shown in \cref{tab: ptrue}, the experimental results indicate that $\hat{p}_{true}$ is superior in responding to the importance of modality and correct for the modality results' inconsistencies compared with other fusion weights, demonstrating the reliability of our $\hat{p}_{true}^m$.

\subsection{Drawbacks of Employing Entropy in Composing Relative Calibration} \label{app: uncertainty}
\begin{figure}[b] 
\vskip 0.1in
\begin{center}
\centerline{\includegraphics[width=0.48\textwidth]{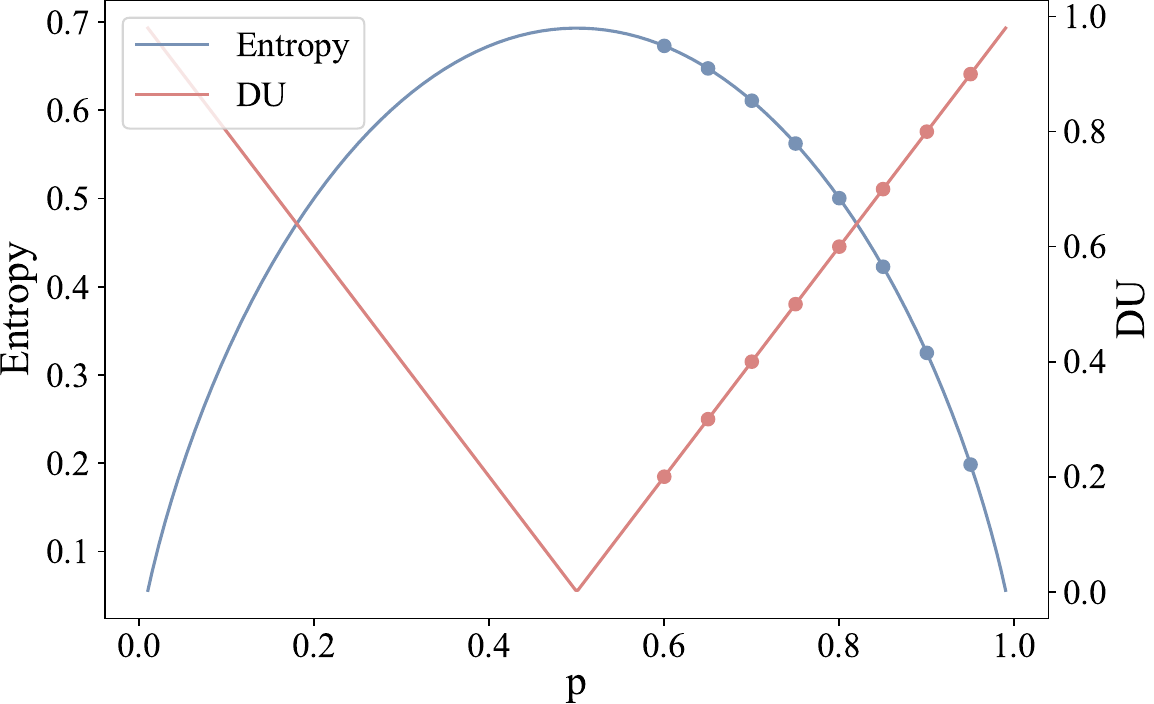}}
\caption{
The function diagram of Entropy and Distribution Uniformity in two categories.} \label{figure: function_GDP}
\end{center}
\vskip -0.1in
\end{figure}
Traditional approaches to evaluating informational uncertainty, exemplified by entropy \cite{shannon1948mathematical}, face difficulties within our relative framework. \cref{figure: function_GDP} (a) demonstrates that when entropy is held constant for a particular modality, the variable rate of change of its derivative engenders erratic fluctuations in the relative value of multimodal entropy in response to variations in another modality's entropy, resulting in biased comparisons. In contrast, Distribution Uniformity (DU), characterized by its constant slope, facilitates more balanced assessments of uncertainty across different modalities.

\subsection{More Comparisons with Other Methods}
The recognition of decision-level generalization bounds in past analyses by QMF is commendable. Nevertheless, it is crucial to acknowledge that our methodology diverges markedly at the conceptual stage from QMF's approach. QMF's formula for generalization bounds, grounded in unimodal analysis, overlooks the complex interplay among modalities within a multimodal system. The terms 'Term-L' and 'Term-C' within QMF's framework are not static, varying with each modality's response to scene changes. Consequently, enforcing a negative $\text{Term-Cov}$ through calibration does not ensure a reduced generalization bound. Our perspective posits that each modality's contribution in a multimodal system should be assessed in relation, and with this view, we have deduced the Generalization Error Bound (GEB) through a multimodal perspective. Fundamentally, the GEB is influenced only by Mono-Term and Holo-Term, considering that optimization within the same function class yields identical Rademacher complexity and empirical error. Therefore, ensuring the sign of Mono-Term and Holo-Term suffices to diminish the generalization bound.

Furthermore, QMF endeavors to gauge the quality of modalities by accounting for uncertainty, postulating a direct relationship between uncertainty and loss, thereby aligning its approach with its theoretical framework of the generalization bound. This necessitates the application of calibration to uphold the initial premise. However, our analysis of the $p_{true}$, which indicates the modality's reliability, reveals an inverse relationship with the loss. Consequently, rooted in the Generalization Error Bound (GEB) concept, we propose a strategy to diminish the GEB by focusing on the accurate prediction of $p_{true}$.

\section{More Details}
\subsection{Related Work Details}
\textbf{Predictive confidence} is currently frequently used in fault detection \cite{corbiere2019addressing} and Out-Of-Distribution (OOD) detection \cite{devries2018learning}. However, it often leads to overconfidence due to softmax probabilities. Some methods focus on smoothing the predicted probability distribution through label smoothing \cite{muller2019does}, while others apply temperature scaling \cite{liang2017enhancing} to calibrate the probability outputs. Some works \cite{wei2022mitigating} mitigate overconfidence by constraining the magnitude of logits. Essentially, most of these approaches aim to align the expected class probabilities with empirical accuracy \cite{ma2023calibrating}.

\subsection{Symbols Table}
To avoid potential confusion, we provide a table for main symbols in \cref{tab:symble_table}.
\begin{table}[t]
\caption{Main Symbols Table.} \label{tab:symble_table}
\vskip 0.1in
\renewcommand\arraystretch{1.1}
\tabcolsep=0.05cm
\begin{center}
\begin{small}
\begin{sc}
\begin{tabular}{cc}
\toprule
  Symbol & Explanation  \\
\midrule
$\mathcal{M}$   &  The set of each uni-modal \\
$\lvert\mathcal{M}\rvert$  &  The cardinality of $\mathcal{M}$  \\
$\omega^m$  &  Fusion weight of the m-th modality \\
$f^m$  &  Uni-modal projection function \\
GEB($f$)  & Generalization Error upper Bound of $f$  \\
$p_{true}$  & True class probability \\
$\hat{p}_{true}$  &  The prediction of $p_{true}$  \\
$\ell$  &  Logistic loss function  \\
$\ell^m$  &  The simplicity representation of $\ell(f^m(x^m),y)$ \\
$\hat{\ell^m}$  &  The prediction of $\ell^m$ \\
$\hat{err}(f^m)$  &  Empirical errors of the m-th modality \\
$\mathcal{H}$  & Hypothesis set  \\
$R_N(\mathcal{H})$  &  Rademacher complexities  \\
\bottomrule
\end{tabular}
\end{sc}
\end{small}
\end{center}
\vskip -0.1in
\end{table}

\subsection{Metric Details}
\label{app:metric}
To quantify the capacity of a fusion strategy on reducing the GEB, we defined a metric called \textbf{G}EB \textbf{D}ecreasing \textbf{P}roportion (GDP):
\begin{align} \label{eq:GDP}
    \text{GDP}=\mathbb E_\mathcal{F}[\mathbb{I}_{\{\text{AC}(f)<0\vert f\in\mathcal{H}\}}(f)],
\end{align}
where $\mathcal{F}\subset\mathcal{H}$, $\mathbb{I}$ is the indicator function, which is defined as:
\begin{align}
    \mathbb{I}_{\{\text{AC}(f)<0\vert f\in\mathcal{H}\}}(f)=\begin{cases}
        1 &\text{if } f\in \{\text{AC}(f)<0 \vert f\in\mathcal{H}\},\\
        0 &\text{otherwise}.
    \end{cases}
\end{align}

\subsection{Implementation Details}\label{app:implement}
The network was trained for 100 epochs utilizing the Adam optimizer with $\beta_1=0.9$, $\beta_2=0.999$, weight decay of 0.01, dropout rate of 0.1, and a batch size of 16. 
The initial learning rate was chosen from the set \{$1e-8$, $5e-5$,  $1e-4$\}. Specifically, for image-text classification, the initial learning rate was $5e-5$; for scene recognition, it was 1e-8 for the second layer of the confidence predictor and 1e-4 for all others; for emotion recognition, it was set to 1e-3. 
All the experiments were conducted on an NVIDIA A6000 GPU, using PyTorch with default parameters for all methods.
\subsection{Experiment Details}\label{sec:noise details}
We changed the data quality by varying noise levels. Specifically, we alter the degree of noise added to it and fix another modality's noise level. For NYU Depth V2 dataset, we fixed the RGB noise level at 5 and increased the depth noise from 0 to 10. For MVSA and FOOD101 datasets, we maintained the text noise at 2.5 and escalated the image noise from 0 to 5. For the CREMA-D dataset, we kept the audio modality's SNR fixed and varied the image noise from 0 to 10. With the increase of one modality's noise, we report the changing trend of each modality's calibrated weight.

\subsection{\texorpdfstring{The Prediction of $p^{m}_{true}$ and Loss Function}{The Prediction of p\^m\_true and Loss Function}}\label{sec:loss}
During the inference phase, there is no ground truth available to get the $p_{true}^m$, so we trained a confidence predictor consisting of multiple linear layers to predict $\hat{p}_{true}^m$ by the MSELoss:
\begin{align} \label{loss:tcp} \mathcal{L}_{{p}_{true}}=\sum_{m=1}^{|\mathcal{M}|}{MSE\left(\hat{p}_{true}^m,p_{true}^m\right)},
\end{align}

where $\hat{p}_{true}^m=\mathrm{Predictor}(feature^m)$, and ${feature^m}$ is the feature of input $x^{m}$ generated by encoder.
Leveraging the ${\hat{p}_{true}}^m$  to compute $\text{Co-Belief}^m$, utilizing the model's output after softmax to calculate $\mathrm{RC}^m$ and its transformation $k_m$, the final calibrated Co-Belief $\text{CCB}^m$ as well as $\omega^m$ can be obtained as \cref{eq:cali}.

Drawing on the principles of multi-task learning, we conceive the overall loss function as the aggregate of standard cross-entropy classification losses across multiple modalities, coupled with the $p_{true}$ prediction loss:
\begin{align}
    \mathcal{L}_{\text{overall}}=\mathcal{L}_{\text{CE}}(y,f(x))+\sum_{m=1}^{|\mathcal{M}|}{\mathcal{L}_{\text{CE}}(y,f^m(x^m))}+\mathcal{L}_{{p}_{true}} \label{loss:all},
\end{align}
where $\mathcal{L}_{\text{CE}}$ represents the cross-entropy loss, while $\mathcal{L}_{{p}_{true}}$ is the $p_{true}$ prediction loss defined as \cref{loss:tcp}. 

\begin{figure}[t]
\begin{center}
\centerline{\includegraphics[width=0.7\textwidth]{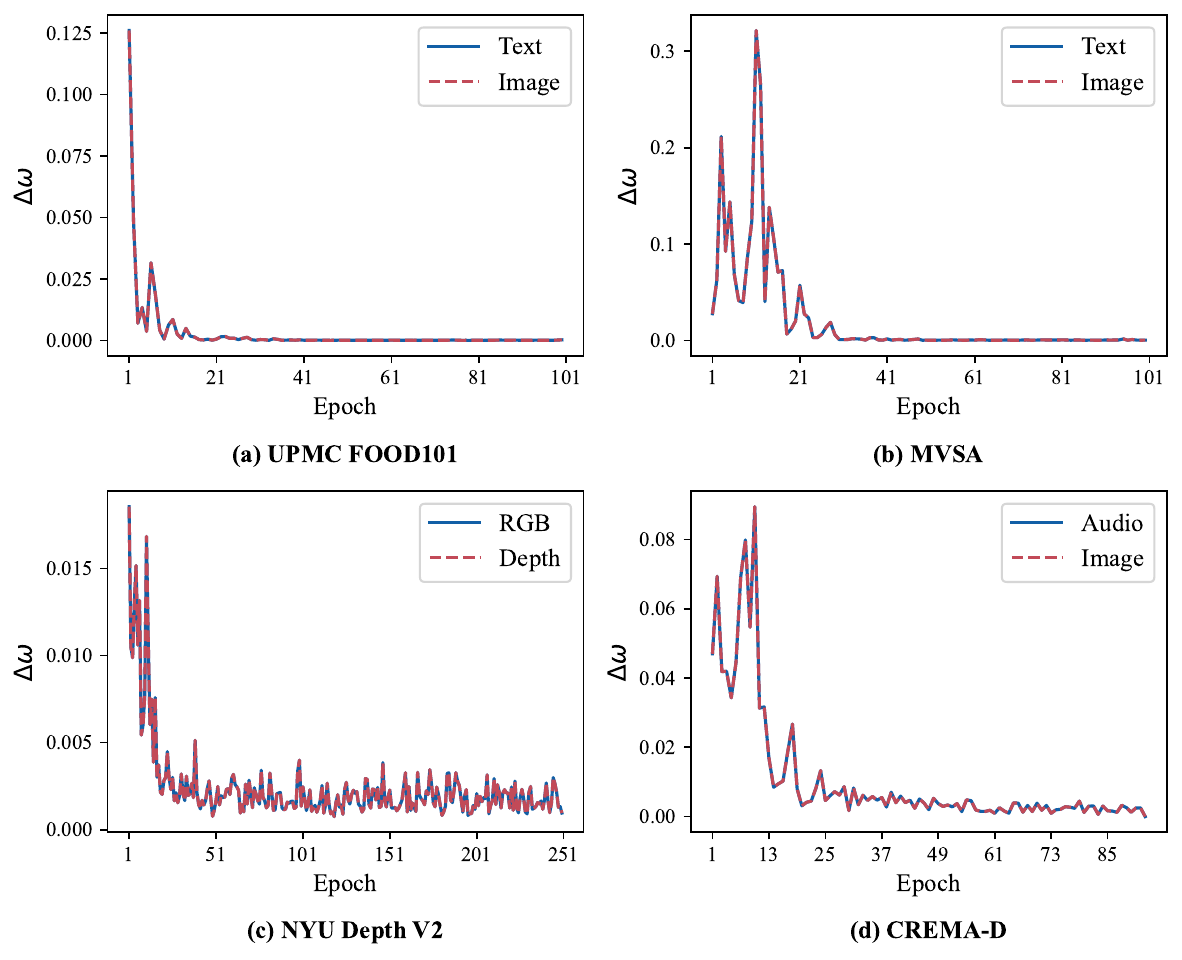}}
\caption{
$\Delta \omega^m$ gradually shrinks to zero as the training epochs increase.} \label{fig:convergence}
\end{center}
\vskip -0.11in
\end{figure}

\begin{table}[t]
\caption{Comparative experiments on asymmetric calibration and its variants with NYU Depth V2. }
\label{sample-table}\label{table:rr}
\vskip 0.1in
\renewcommand\arraystretch{1.2}
\begin{center}
\begin{small}
\begin{sc}
\begin{tabular}{ccll}
\toprule
variants of RC & $\epsilon$ = $0$ & $\epsilon$ = $5$ & $\epsilon$ = $10$ \\ \midrule
Distribution Uniformity (DU)       & $71.13$          & $64.93$          & $62.08$          \\
Relative Calibration (RC)    & $71.06$          & $65.43$          & $62.19$          \\
asymmetric calibration          & $\mathbf{71.37}$ & $\mathbf{65.72}$ & $\mathbf{62.56}$  \\ \bottomrule
\end{tabular}
\end{sc}
\end{small}
\end{center}
\vskip -0.1in
\end{table}
\section{Additional results}
\subsection{Lower Generalization Error Upper Bound} \label{app:lower GEB}
We applied the proposed Generalization Error Bound Decreasing Proportion (GDP, as defined in \cref{eq:GDP}) to measure the capacity of a fusion strategy for reducing the Generalization Error Upper Bound (GEB). Separately, we compare the GDP metrics of models that utilize ${\text{Mono-Conf}}$ (\cref{equ:w1}), ${\text{Co-Belief}}$ (\cref{equ:inter}), and $\text{CCB}$ (\cref{eq:cali}) as fusion weights.
Specifically, for each fusion strategy, we trained 50 models with distinct random seeds and reported the GDP of these models under varying noises (0, 5, and 10) during testing.
We took Mono-Confidence, Co-Belief, and CCB as distinct fusion weights for the model and depicted the AC distribution in \cref{figure: GDP} (a), (b), and (c). The proportion of $\text{AC} < 0$, i.e. GDP, signifies the fusion strategy's potential to reduce the model's GEB. 
\cref{figure: GDP} (d) displays the GDP of different fusion strategies under diverse noisy conditions. The proportion of Mono-Confidence that reduces GEB without noise is greater than 50\%, and Holo-Confidence further increases the model's GDP, CCB endows the model with the best generalization capabilities in dynamic environments. These experiments validate our effectiveness in lowering the generalization error upper bound.

On comparing \cref{figure: GDP} (a) and (b), it is evident that the extent of the orange segment in \cref{figure: GDP} (b), i.e., the model's GDP, is greater than that in \cref{figure: GDP} (a), suggesting that the integration of both Mono-Confidence and Holo-Confidence yields a heightened likelihood of GEB reduction. In \cref{figure: GDP}  (c), the application of Calibrated Co-Belief as the fusion weight demonstrates a probability of up to 90\% in reducing the generalization error.

\cref{figure: GDP} (d) displays the GDP of different fusion strategies under various noisy conditions, with the sum corresponding to the GDP of the three fusion strategies depicted in \cref{figure: GDP} (a) (b) and (c). We observe that the proportion of Mono-Confidence reducing GEB without noise is greater than 50\%, and the inclusion of Holo-Confidence further increases the model's GDP, indicating an enhanced generalization ability. Additionally, the incorporation of the calibration strategy endows the model with stronger generalization capabilities in noisy environments.
\subsection{Convergence of Fusion Weight}\label{app:convergence}
To demonstrate the dependability of our calibrated weights, we ascertain their convergence when training.
We define $\Delta\omega^m$ as the mean absolute change of the weights across the entire validation set for each epoch and track its progression throughout the training period. As illustrated in \cref{fig:convergence}, across various datasets, $\Delta\omega^m$ consistently trends toward zero, indicating the calibrated fusion weights' convergence.

\subsection{Effectiveness of the Form of the Asymmetric Calibration} \label{app:RR}
We adopt an asymmetric form for Relative Calibration (RC), which not only aligns with the motivation of calibration but also facilitates better weight optimization for the RC with an asymmetric form ranging from 0 and 1.
To validate the effectiveness of asymmetric calibration, we conducted comparative experiments involving asymmetric calibration and its variants, including DU (\cref{equ:DB}), and RC (\cref{equ:grb}) without asymmetric form by assessing their impact on average accuracy metrics. Results presented in \cref{table:rr} suggest that RC with asymmetric calibration exhibits enhanced generalization capabilities. 

\subsection{Extensibility to Datasets with More than 2 Modalities} \label{app: multimodal}
As illustrated in \cref{eq:multimodal}, our proposed relative calibration can be extended to cases where $\lvert\mathcal{M}\rvert>2$. To verify the effectiveness of our PDF, we conducted comparisons with previous state-of-the-art methods on the PIE dataset with three modalities. As depicted in \cref{tab:pie}, our PDF surpasses the competing methods across various noise levels, highlighting its superiority.

\begin{table}[t] 
\caption{Comparison on three modalities dataset PIE under Gaussian noise.}\label{tab:pie}
\vskip 0.1in

\begin{center}

\begin{small}
\begin{sc}
\centering
\tabcolsep=0.4cm

\begin{tabular}{lcccccc}

\toprule
Method    & $\epsilon=0.0$ & $\epsilon=2.0$ & $\epsilon=4.0$ & $\epsilon=6.0$ & $\epsilon=8.0$ & $\epsilon=10.0$ \\ 
\midrule
Late Fusion & $88.24$ & $85.74$ & $83.09$ & $81.32$ & $78.09$ & $74.12
$ \\
TMC & $89.71$  & $84.26$ & $79.12$  & $74.71$ & $70.88$  & $64.85$  \\
QMF & $88.24$ & $85.59$ & $81.76$ & $81.03$ & $77.94$ & $74.00$\\
DynMM & $89.71$ & $85.29$ &$84.56$&$82.35$&$80.15$&$76.47$\\
\midrule
Ours & $\mathbf{90.44}$ & $\mathbf{88.82}$ & $\mathbf{84.71}$ & $\mathbf{82.50}$ & $\mathbf{80.74}$ & $\mathbf{78.53}$\\
\bottomrule
\end{tabular}
\vskip -0.1in
\end{sc}
\end{small}
\end{center}
\end{table}

\subsection{Compared Experiments on Different Noises} \label{app:full_noise}
In this section, we report the full experiment results with standard deviation in varying Gaussian noise and Salt-pepper noise compared with other methods in \cref{tab: Gaussian} and \cref{table: salt} separately.  
\section{Limitations}
Even though the proposed PDF model achieves superior performance over existing methods and shows advanced generalization ability in dynamically changing conditions, there are still some potential limitations. We provide theoretical guarantees to the Co-Belief prediction, however, the potential uncertainty is inevitable. The proposed relative calibration is an empirical solution for this problem without theoretical guarantees. Therefore, it is important to explore new uncertainty estimation methods from the theoretical perspective. Besides, the predictor in our model is relatively simple, and it is also valuable to study more efficient and effective predictor architectures.
\begin{table*}[t]
\renewcommand\arraystretch{1.4}
\caption{We add Gaussian noise on $50\%$ modalities and $\epsilon$ presents the noise degree. it shows the average and the standard deviation of classification accuracies with our method and the compared methods on four datasets. The method marked with * was replicated by us, while the rest of the data is sourced from \cite{10.5555/3618408.3620161}.} \label{tab: Gaussian}
\vskip 0.10in
\begin{center}
\begin{sc}
\tabcolsep=0.5cm
\begin{tabular}{c|cccc}
\toprule
Dataset                                                                   & Method     & $\epsilon$ = $0.0$                                  &  $\epsilon$ = $5.0$                                  &  $\epsilon$ = $10.0$                                 \\ \midrule
                                                                          & Text        & $75.61$$\pm$$0.53$                                 & $69.50$$\pm$$1.50$& $47.41$$\pm$$0.79 $                                \\
                                                                          & img        & $64.12$$\pm$$1.23 $                                & $49.36$$\pm$$2.02$                                 & $45.00$$\pm$$2.63$\\
                                                                          & concat & $65.59$$\pm$$1.33$                                 & $50.70$$\pm$$2.65$& $46.12$$\pm$$2.44$                                 \\
                                                                          & late fusion & $76.88$$\pm$$1.30$& $63.46$$\pm$$3.46$                                 & $55.16$$\pm$$3.60$\\
                                                                          & QMF        & $78.07$$\pm$$1.10$& $73.85$$\pm$$1.42 $                                & $61.28$$\pm$$2.12$                                 \\
                                                                          & TMC        & $74.87$$\pm$$2.24 $                                & $66.72$$\pm$$4.55 $                                & $60.35$$\pm$$2.79$                                 \\
                                                                          & dynMM*      & $79.07$$\pm$$0.53$                                 & $67.96$$\pm$$1.65$                                 & $59.21$$\pm$$1.41$                                 \\ \cline{2-5} 
\multirow{-8}{*}{MVSA}                                          \rule{0pt}{11pt}   & Ours       & {\color[HTML]{000000} $\mathbf{79.94}$$\bm{\pm}$$\mathbf{0.95}$} & {\color[HTML]{000000} $\mathbf{74.4}$$\bm{\pm}$$\mathbf{1.51}$}  & {\color[HTML]{000000} $\mathbf{63.09}$$\bm{\pm}$$\mathbf{1.33}$} \\ \midrule
                                                                          & text        & $86.46$$\pm$$0.05$                                 & $67.38$$\pm$$0.19$                                 & $43.88$$\pm$$0.32$                                 \\
                                                                          & img        & $64.62$$\pm$$0.40$& $34.72$$\pm$$0.53$                                 & $33.03$$\pm$$0.37$                                 \\
                                                                          & concat & $88.20$$\pm$$0.34$& $61.10$$\pm$$2.02$& $49.86$$\pm$$2.05$                                 \\
                                                                          & late fusion & $90.69$$\pm$$0.12$                                 & $68.49$$\pm$$3.37$                                 & $57.99$$\pm$$1.59$                                 \\
                                                                          & QMF        & $92.92$$\pm$$0.11$                                 & $76.03$$\pm$$0.70$& $62.21$$\pm$$0.25$                                 \\
                                                                          & TMC        & $89.86$$\pm$$0.07$                                 & $73.93$$\pm$$0.34$                                 & $61.37$$\pm$$0.21$                                 \\
                                                                          & dynMM*      & $92.59$$\pm$$0.07 $                                & $74.74$$\pm$$0.19 $                                & $59.68$$\pm$$0.20$\\ \cline{2-5} 
\multirow{-8}{*}{\begin{tabular}[c]{@{}c@{}}UMPC \\FOOD 101\end{tabular}}                                                \rule{0pt}{11pt}  & Ours       & {\color[HTML]{000000} $\mathbf{93.32}$$\bm{\pm}$$\mathbf{0.22}$} & {\color[HTML]{000000} $\mathbf{76.47}$$\bm{\pm}$$\mathbf{0.31}$ } & {\color[HTML]{000000} $\mathbf{62.83}$$\bm{\pm}$$\mathbf{0.31}$} \\ \midrule
                                                                          & RGB        & $62.65$$\pm$$1.22$                                 & $50.95$$\pm$$3.38$                                 & $44.13$$\pm$$3.80$\\
                                                                          & Depth      & $63.30$$\pm$$0.48$& $53.12$$\pm$$1.52$                                 & $45.46$$\pm$$2.07$                                 \\
                                                                          & concat*     & $69.88$$\pm$$0.52$                                 & $63.82$$\pm$$1.46$                                 & $60.03$$\pm$$2.63$                                 \\
                                                                          & late fusion* & $70.03$$\pm$$0.84$                                 & $64.37$$\pm$$0.80$& $60.55$$\pm$$1.65$                                 \\
                                                                          & TMC*        & $70.40$$\pm$$0.31$& $59.33$$\pm$$2.19$                                 & $50.61$$\pm$$2.87$                                 \\
                                                                          & QMF*        & $69.54$$\pm$$1.06$                                 & $64.10$$\pm$$1.42$& $60.18$$\pm$$1.23 $                                \\
                                                                          & DynMM*      & $65.50$$\pm$$0.37$& $54.31$$\pm$$1.72$                                 & $46.79$$\pm$$1.09$                                 \\
                                                                           \cline{2-5} 
\multirow{-9}{*}{\begin{tabular}[c]{@{}c@{}}NYU \\ Depth V2\end{tabular}} \rule{0pt}{11pt}  & Ours       & {\color[HTML]{000000} $\mathbf{71.37}$$\bm{\pm}$$\mathbf{0.76}$} & {\color[HTML]{000000} $\mathbf{65.72}$$\bm{\pm}$$\mathbf{1.72}$} & {\color[HTML]{000000} $\mathbf{62.56}$$\bm{\pm}$$\mathbf{1.84}$} \\ \midrule
                                                                          & visual*     & $43.60$$\pm$$2.02$& $32.52$$\pm$$1.98$                                 & $30.17$$\pm$$1.19$                                 \\
                                                                          & audio*      & $58.67$$\pm$$1.01$                                 & $54.66$$\pm$$2.16$                                 & $43.01$$\pm$$6.04$                                 \\
                                                                          & concat*     & $61.56$$\pm$$1.37$                                 & $52.33$$\pm$$3.32$                                 & $41.01$$\pm$$5.70$\\
                                                                          & late fusion* & $61.81$$\pm$$2.13 $                                & $49.84$$\pm$$3.72$                                 & $39.15$$\pm$$5.82$                                 \\
                                                                          & QMF*        & $63.04$$\pm$$1.37$                                 & $56.60$$\pm$$2.38$& $41.60$$\pm$$2.75$\\
                                                                          & TMC*        & $59.15$$\pm$$1.95$                                 & $54.42$$\pm$$3.34$                                 & $46.79$$\pm$$4.72$                                 \\
                                                                          & dynMM*      & $60.46$$\pm$$0.37$                                 & $54.43$$\pm$$0.73$                                 & $42.39$$\pm$$0.50$\\ \cline{2-5} 
\multirow{-8}{*}{CREMA-D}                                               \rule{0pt}{11pt}   & Ours       & {\color[HTML]{000000} $\mathbf{63.31}$$\bm{\pm}$$\mathbf{1.11}$} & {\color[HTML]{000000} $\mathbf{57.85}$$\bm{\pm}$$\mathbf{2.04}$} & {\color[HTML]{000000} $\mathbf{47.84}$$\bm{\pm}$$\mathbf{2.32}$} \\ \bottomrule
\end{tabular}
\end{sc}
\end{center}
\vskip -0.1in
\vspace{-1em}
\end{table*}

\begin{table*}[t]
\renewcommand\arraystretch{1.4}
\caption{We add Salt-pepper noise on $50\%$ modalities and $\epsilon$ presents the noise degree. it shows the average and the standard deviation of classification accuracies with our method and the compared methods on four datasets. The method marked with * was replicated by us, while the rest of the data is sourced from \cite{10.5555/3618408.3620161}.}
\label{table: salt}
\vskip 0.10in
\begin{center}
\begin{sc}
\tabcolsep=0.5cm
\begin{tabular}{c|cccc}
\toprule
\multicolumn{1}{l|}{dataset}                                             & Method       & $\epsilon$ = 0.0                                  & $\epsilon$ = 5.0                                  & $\epsilon$ = 10.0                                 \\ \midrule
                                                                         & text         & $75.61$$\pm$$0.53$                                 & $69.50$$\pm$$1.50$                                 & $47.41$$\pm$$0.79$                                 \\
                                                                         & img          & $64.12$$\pm$$1.23$                                 & $56.72$$\pm$$1.92$                                 & 50.71$\pm$3.20                                 \\
                                                                         & concat       & $65.59$$\pm$$1.33$                                 & $58.69$$\pm$$2.25$                                 & $51.16$$\pm$$2.99$                                \\
                                                                         & late fusion  & $76.88$$\pm$$1.30$                                 & $67.88$$\pm$$1.87$                                 & $55.43$$\pm$$1.94$                                \\
                                                                         & QMF          & $78.07$$\pm$$1.10$                                & $73.90$$\pm$$1.89$                                 & $60.41$$\pm$$2.63$                                 \\
                                                                         & TMC          & $74.87$$\pm$$2.24$                                 &$ 68.02$$\pm$$3.07$                                 & 56.62$\pm$3.67                                 \\
                                                                         & dynMM*       & $79.07$$\pm$$0.53$                                 & $71.35$$\pm$$0.97$                                 & $59.96$$\pm$$1.31$                                 \\ \cline{2-5} 
\multirow{-8}{*}{MVSA}                                           \rule{0pt}{11pt}  & Ours         & {\color[HTML]{000000} {$\mathbf{79.94}$$\bm\pm$$\mathbf{0.95}$}} & {\color[HTML]{000000} $\mathbf{75.11}$$\bm\pm$$\mathbf{1.15}$} & {\color[HTML]{000000} $\mathbf{61.97}$$\bm\pm$$\mathbf{1.14}$} \\ \midrule
                                                                         & text         & $86.44$$\pm$$0.02$                                 & $67.41$$\pm$$0.20$                                 & $43.89$$\pm$$0.33$                                 \\
                                                                         & img          & $64.53$$\pm$$0.47$                                 & $50.75$$\pm$$0.44 $                                & $36.83$$\pm$$0.92$                                 \\
                                                                         & concat       & $88.22$$\pm$$0.36$                                 & $72.49$$\pm$$0.75$                                 & $52.10$$\pm$$0.97$                                 \\
                                                                         & late fusion  & $90.66$$\pm$$0.16$                                 & $77.99$$\pm$$0.54$                                 & $58.75$$\pm$$0.99$                                 \\
                                                                         & QMF          & $92.90$$\pm$$0.13$                                 &$ 80.87$$\pm$$0.40$                                 & $61.60$$\pm$$0.20$                                 \\
                                                                         & TMC          & $89.86$$\pm$$0.07$                                 & $77.86$$\pm$$0.41$                                 & $60.22$$\pm$$0.43$                                 \\
                                                                         & dynMM*       & $92.59$$\pm$$0.07$                                 & $78.91$$\pm$$0.20$                                 & $57.64$$\pm$$0.30$                                 \\ \cline{2-5} 
\multirow{-8}{*}{\begin{tabular}[c]{@{}c@{}}UMPC\\ FOOD101\end{tabular}} \rule{0pt}{11pt} & Ours         & {\color[HTML]{000000} $\mathbf{93.32}$$\bm\pm$$\mathbf{0.22}$} & {\color[HTML]{000000} $\mathbf{81.21}$$\bm\pm$$\mathbf{0.34}$} & {\color[HTML]{000000} $\mathbf{61.76}$$\bm\pm$$\mathbf{0.33}$} \\ \midrule
                                                                         & RGB          & $62.61$$\pm$$1.21$                                 & $49.14$$\pm$$1.40$                                 & $34.76$$\pm$$1.59 $                                \\
                                                                         & Depth        & $63.32$$\pm$$0.50$                                 & $50.99$$\pm$$1.41$                                 & 38.56$\pm$2.16                                 \\
                                                                         & concat*      & $69.88$$\pm$$0.52$                                 & $61.41$$\pm$$1.69$                                 & $51.65$$\pm$$2.94$                                 \\
                                                                         & late fusion* & $70.03$$\pm$$0.84$                                 & $62.05$$\pm$$1.17$                                 & $51.50$$\pm$$1.81 $                                \\
                                                                         & TMC*         & $70.40$$\pm$$0.31$                                 & $59.33$$\pm$$1.47$                                 & $45.32$$\pm$$2.84$                                 \\
                                                                         & QMF*         & $69.54$$\pm$$1.06$                                 & $62.02$$\pm$$1.47$                                 & $51.87$$\pm$$0.91$                                 \\
                                                                         & DynMM*       &$ 65.50$$\pm$$0.37 $                                & $52.26$$\pm$$1.45 $                                & $38.17$$\pm$$1.17 $                                \\
                                                                          \cline{2-5} 
\multirow{-9}{*}{\begin{tabular}[c]{@{}c@{}}NYU\\ Depth V2\end{tabular}}\rule{0pt}{11pt}  & Ours         & {\color[HTML]{000000} $\mathbf{71.37}$$\bm\pm$$\mathbf{0.76}$} & {\color[HTML]{000000} $\mathbf{64.27}$$\bm\pm$$\mathbf{1.36}$} & {\color[HTML]{000000} $\mathbf{53.62}$$\bm\pm$$\mathbf{2.15}$} \\ \midrule
                                                                         & visual*      & $43.60$$\pm$$2.02$                                 & $40.30$$\pm$$1.77$                                 & $36.84$$\pm$$1.72$                                 \\
                                                                         & audio*       & $58.67$$\pm$$1.01$                                 & $54.57$$\pm$$2.06$                                 & $43.00$$\pm$$6.01$                                 \\
                                                                         & concat*      & $61.56$$\pm$$1.37$                                 & $54.28$$\pm$$3.89$                                 & $42.57$$\pm$$6.16$                                 \\
                                                                         & late fusion* & $61.81$$\pm$$2.13$                                 & $54.83$$\pm$$3.24$                                 & $41.07$$\pm$$6.88$                                 \\
                                                                         & QMF*         & $63.04$$\pm$$1.37$                                 &$ 57.73$$\pm$$2.25$                                 & $45.02$$\pm$$2.28$                                 \\
                                                                         & TMC*         & $59.15$$\pm$$1.95$                                 & $54.61$$\pm$$3.19$                                 & $47.72$$\pm$$2.76$                                 \\
                                                                         & dynMM*       & $60.46$$\pm$$0.37$                                 &$ 54.58$$\pm$$0.65$                                 & $42.49$$\pm$$0.43$                                 \\ \cline{2-5} 
\multirow{-8}{*}{Cremad}                                                \rule{0pt}{11pt}  & Ours         & {\color[HTML]{000000} $\mathbf{63.31}$$\bm\pm$$\mathbf{1.11}$} & {\color[HTML]{000000} $\mathbf{58.61}$$\bm\pm$$\mathbf{1.50}$}& {\color[HTML]{000000} $\mathbf{48.40}$$\bm\pm$$\mathbf{2.85}$}  \\ \bottomrule
\end{tabular}
\end{sc}
\end{center}
\vskip -0.1in
\vspace{-1em}
\end{table*}

\end{document}